%% file: main.tex
\documentclass[10pt,journal,compsoc]{IEEEtran}
\ifCLASSOPTIONcompsoc
  \usepackage[nocompress]{cite}
\else
  \usepackage{cite}
\fi
\ifCLASSINFOpdf
    \usepackage{subcaption}
    \usepackage{multirow}
    \usepackage{adjustbox}
    \usepackage{wrapfig}
    \usepackage{pgfplots}
    \usepackage{colortbl}
    \usepgfplotslibrary{groupplots, colormaps}
    \pgfplotsset{compat=1.17} 
\else
\fi
\usepackage{amsmath}
\usepackage{times}    
\usepackage{xspace}
\usepackage{graphicx}
\usepackage{amssymb}
\usepackage{booktabs}
\usepackage{enumitem}
\begin{document}
%
\title{AI-SAM: Automatic and Interactive Segment Anything Model}
%
%
%
%

\author{Yimu~Pan, Sitao~Zhang, Alison~D.~Gernand, Jeffery~A.~Goldstein, and James~Z.~Wang,~\IEEEmembership{Senior~Member,~IEEE}
\IEEEcompsocitemizethanks{\IEEEcompsocthanksitem Y. Pan, S. Zhang, and J. Z. Wang are with the Data Science and Artificial Intelligence Area, College of Information Sciences and Technology, The Pennsylvania State University, University Park, PA, 16802, USA. J. Z. Wang is also with the Huck Institutes of the Life Sciences, The Pennsylvania State University, University Park, PA, 16802, USA. Correspondence should be addressed to Y. Pan.\protect\\
E-mails: ymp5078@psu.edu, sitao.zhang@psu.edu, jwang@ist.psu.edu
\IEEEcompsocthanksitem A. D. Gernand is with the Department of Nutritional Sciences, College of Health and Human Development, The Pennsylvania State University, University Park, PA, 16802, USA.\protect\\
Email: adg14@psu.edu
\IEEEcompsocthanksitem J. A. Goldstein is with the Department of Pathology, Northwestern Memorial Hospital, Northwestern University, Chicago, IL, 60611, USA.\protect\\
Email: ja.goldstein@northwestern.edu
\IEEEcompsocthanksitem Research reported in this publication was supported by the National Institute of Biomedical Imaging and Bioengineering of the National Institutes of Health (NIH) under award number R01EB030130. The content is solely the responsibility of the authors and does not necessarily represent the official views of the NIH. This work used computing resources at the National Center for Supercomputing Applications through allocation IRI180002 from the Advanced Cyberinfrastructure Coordination Ecosystem: Services \& Support (ACCESS) program, which is supported by National Science Foundation grants Nos. 2138259, 2138286, 2138307, 2137603, and 2138296.
}
\thanks{Manuscript received December 10, 2023; revised .}}

\IEEEtitleabstractindextext{%
\input{sec/0_abstract}

}

\maketitle

\IEEEdisplaynontitleabstractindextext

%
\IEEEpeerreviewmaketitle

\input{sec/1_intro}
\input{sec/2_relatedworks}
\input{sec/3_method}

\input{sec/4_results}

\appendices

\input{sec/X_suppl}

\ifCLASSOPTIONcompsoc
\else
\fi


\ifCLASSOPTIONcaptionsoff
  \newpage
\fi



\bibliographystyle{IEEEtran}
\bibliography{IEEEabrv,main}

%

\begin{IEEEbiography}[{\includegraphics[width=0.9in,height=1.25in,clip,keepaspectratio,trim=0 0 0 0]{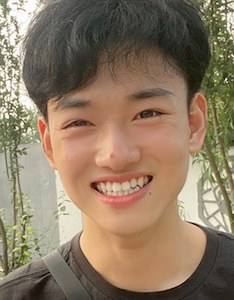}}]{Yimu Pan} is a Ph.D. candidate in the Informatics program at The Pennsylvania State University, advised by James Z. Wang. His research centers around the fields of computer vision and machine learning, with a specific emphasis on vision-language techniques and medical image applications. Before starting his Ph.D. program, he received bachelor's degrees in Computer Science, Mathematics, and Statistics from The Pennsylvania State University.
\end{IEEEbiography}

\begin{IEEEbiography}[{\includegraphics[width=0.9in,height=1.25in,clip,keepaspectratio,trim=0 0 0 0]{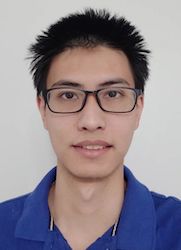}}]{Sitao Zhang} is a Ph.D. candidate in the Informatics program at The Pennsylvania State University, advised by James Z. Wang. His research centers around the fields of computer vision and machine learning, with a specific emphasis on vision-language integration and self-supervised learning. Before joining Penn State, he received a master's degree in Mathematics from the University of Wisconsin-Madison and a bachelor's degree in Statistics from Sun Yat-sen University.

\end{IEEEbiography}

\begin{IEEEbiography}[{\includegraphics[width=0.9in,height=1.25in,clip,keepaspectratio,trim=0 0 0 0]{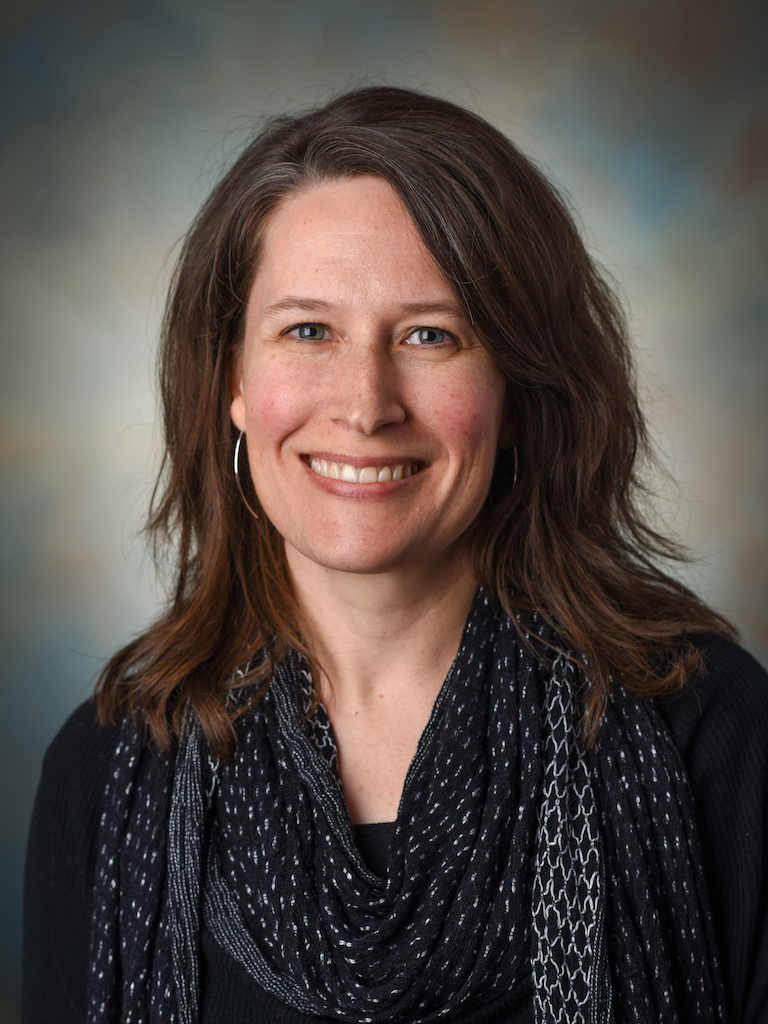}}]{Alison D. Gernand} is an Associate Professor in the Department of Nutritional Sciences at the Pennsylvania State University. She completed her PhD in international health and human nutrition at the Johns Hopkins Bloomberg School of Public Health and her postdoctoral training in nutritional epidemiology at the University of Pittsburgh School of Public Health. Her research is focused on maternal nutritional status and adverse pregnancy outcomes in low- and middle-income countries. Work in low-resource settings has led her to develop and validate simple, low-cost methods to make assessments of the placenta and plasma volume expansion. Her ongoing work leading a large, international collaborative team is developing PlacentaVision, an AI-based software to assess placentas via digital photograph.
\end{IEEEbiography}

\begin{IEEEbiography}[{\includegraphics[width=0.9in,height=1.25in,clip,keepaspectratio,trim=0 0 0 0]{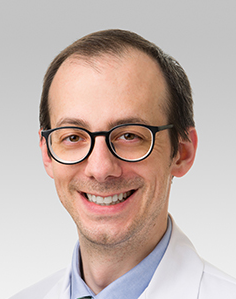}}]{Jeffery A. Goldstein} is the Director of Perinatal Pathology in the Department of Pathology
at Northwestern Memorial Hospital, Northwestern University, Chicago. His long-term goal is to build highly collaborative groups to improve maternal-child health using AI in the placenta. He completed an MD and PhD at the University of Chicago. He completed a residency and fellowship in Anatomic and Pediatric Pathology, with ongoing clinical work in placental pathology. He has worked successfully with teams at the interface of clinical medicine and translational science, as in his work with the AI-PLAX rapid placental diagnostics project, the SNIPP study of COVID-19, and AI identification of gestational age, macroscopic placental lesions, and prostate cancer.
\end{IEEEbiography}



\begin{IEEEbiography}[{\includegraphics[width=0.9in,height=1.25in,clip,keepaspectratio,trim=1 12 0 0]{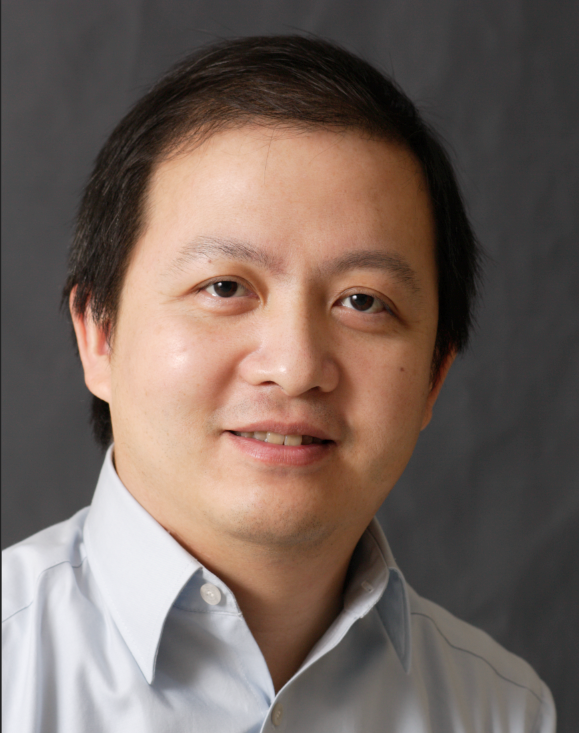}}]
{James Z. Wang} (Senior Member, IEEE)
is a Distinguished Professor of the Data Sciences and Artificial Intelligence section of the College of Information Sciences and Technology at The Pennsylvania State University. He received the bachelor's degree in mathematics {\it summa cum laude} from the University of Minnesota (1994), and the MS degree in mathematics (1997), the MS degree in computer science (1997), and the PhD degree in medical information sciences (2000), all from Stanford University. His research interests include image analysis, affective computing, image modeling, image retrieval, and their applications. He was a visiting professor at the Robotics Institute at Carnegie Mellon University (2007-2008), a lead special section guest editor of the IEEE Transactions on Pattern Analysis and Machine Intelligence (2008), a program manager at the Office of the Director of the National Science Foundation (2011-2012), and a special issue guest editor of the IEEE BITS -- The Information Theory Magazine (2022). He is also affiliated with the Department of Communication and Media, School of Social Sciences and Humanities, Loughborough University, UK (2023-2024). He was a recipient of a National Science Foundation Career Award (2004) and Amazon Research Awards (2018-2022).
\end{IEEEbiography}



\end{document}

%% file: sec/0_abstract.tex
\begin{abstract}

Semantic segmentation is a core task in computer vision. Existing methods are generally divided into two categories: automatic and interactive. Interactive approaches, exemplified by the Segment Anything Model (SAM), have shown promise as pre-trained models. However, current adaptation strategies for these models tend to lean towards either automatic or interactive approaches. Interactive methods depend on prompts user input to operate, while automatic ones bypass the interactive promptability entirely. Addressing these limitations, we introduce a novel paradigm and its first model: the Automatic and Interactive Segment Anything Model (AI-SAM). In this paradigm, we conduct a comprehensive analysis of prompt quality and introduce the pioneering Automatic and Interactive Prompter (AI-Prompter) that automatically generates initial point prompts while accepting additional user inputs. Our experimental results demonstrate AI-SAM's effectiveness in the automatic setting, achieving state-of-the-art performance. Significantly, it offers the flexibility to incorporate additional user prompts, thereby further enhancing its performance. The project page is available at https://github.com/ymp5078/AI-SAM.
\end{abstract}

%% file: sec/1_intro.tex
\IEEEraisesectionheading{\section{Introduction}\label{sec:intro}}


\begin{figure}[t!]
    \centering
    \includegraphics[width=\linewidth]{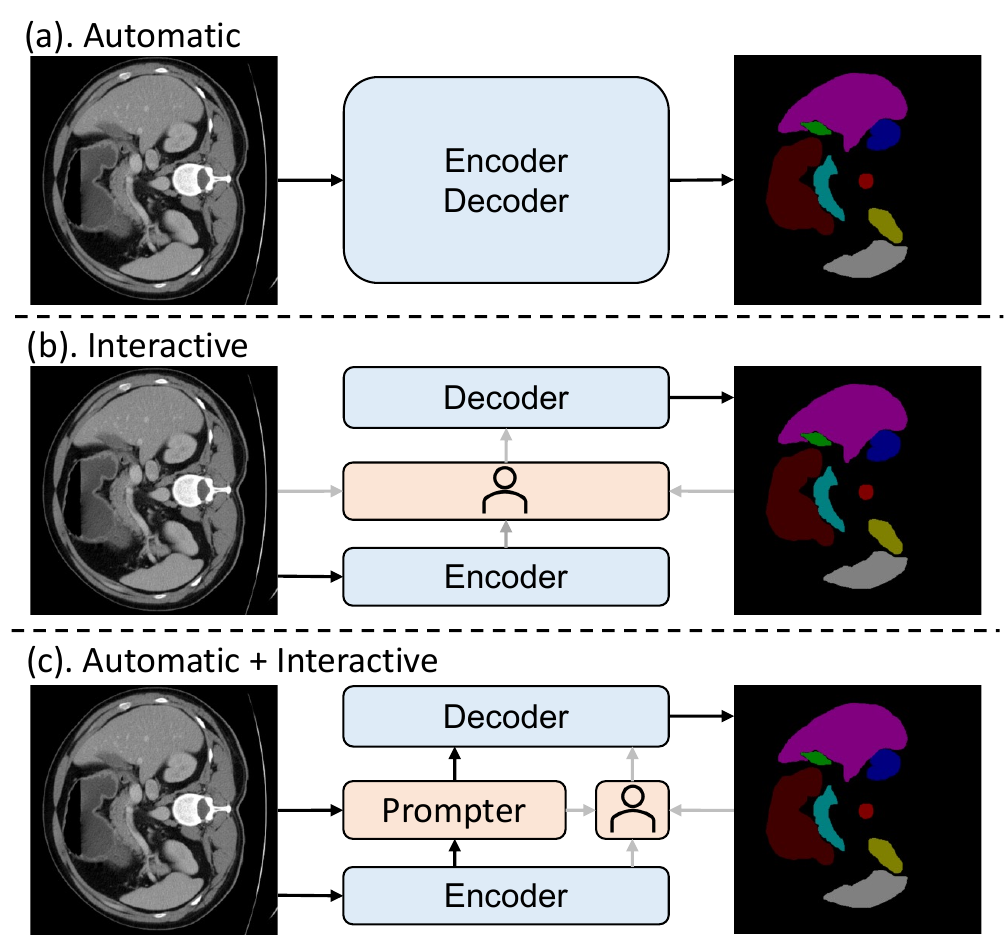}
    \caption{The comparison of three approaches: (a) the automatic approach, (b) the interactive approach, and (c) a hybrid approach that combines elements of both automatic and interactive approaches. Black arrows represent the automatic data flow, whereas the grey arrows represent human intervention.}
    \label{fig:sam_overview}
\end{figure}
\IEEEPARstart{P}{re-trained} foundation models have gained prominence in various domains, owing to their effectiveness in tasks with limited annotations. In the field of image segmentation, interactive segmentation models have garnered attention due to their dual roles: as pre-training models and as data annotation tools. A prime example of this category is the Segment Anything model (SAM)~\cite{kirillov2023segany}, which is acclaimed for its interactive capabilities. It is trained on the large-scale SA-1B dataset, which includes labels based on input prompts (e.g., points, bounding boxes, and text) along with their associated output segmentation masks. Recent studies~\cite{qiu2023large,ma2023segment,ge2023expressive} have demonstrated the transferability of such models to diverse tasks. However, a limitation of SAMs is their inadequate understanding of class semantic granularity, often resulting in subpar performance for tasks with significant domain shifts, unless prior adaptation is conducted. Further, they are trained with ambiguous targets where a single prompt may yield multiple masks, and the resultant masks may not correspond to downstream task semantics (i.e., varying class semantic granularity). To address these issues, some researchers have attempted to delineate semantic granularity by fine-tuning SAM on downstream datasets using synthetic prompts, a process termed `interactive adaptation.' Others have used SAM merely as an initialization step for automatic semantic segmentation tasks.

However, these approaches often compromise SAM's intrinsic capabilities by failing to achieve an optimal balance between automatic and interactive adaptation. Interactive adaptation may lead to underperformance in fully automatic contexts, necessitating human intervention for finalizing the segmentation map regardless of task complexity. Furthermore, while there are numerous possible prompts for interactive methods, their differences have not been comprehensively studied. On the other hand, automatic adaptation completely eliminates SAM's original promptability, rendering the adapted model unsuitable for interactive annotation or intervention during segmentation. Notably, SAM's advantages are most pronounced in limited or bias data scenarios, where iterative model refinement is common. In limited data scenario, engineers typically label additional data in response to a model's ongoing performance. An integrated model that combines both automatic and interactive capabilities could significantly expedite this process, autonomously labeling simpler samples while gradually reducing the need for human input in more complex cases. In bias data scenarios such as~\cite{chen2020ai,zhang2020multi}, the user intervention can be critical for generating correct result. The distinctions between the three approaches are illustrated in Fig.~\ref{fig:sam_overview}

In light of the limitations of current SAM adaptation methods and the potential of a hybrid automatic-interactive model, key research questions arise: (i) How can a model seamlessly combine both automatic and interactive capabilities? (ii) What constitutes an effective prompt for automatic and interactive models? 

To tackle these questions, we introduce a novel automatic and interactive segmentation paradigm. Within this framework, we analyze the characteristics of an effective prompt and propose a new adaptation method, the Automatic and Interactive Segment Anything Model (AI-SAM), which not only preserves the inherent promptability of SAM but also achieves state-of-the-art (SOTA) performance in downstream automatic and interactive segmentation tasks. AI-SAM is designed to complement and enhance current SAM-based automatic adaptation models. Upon generating an initial automatic segmentation result, AI-SAM can incorporate additional user feedback (e.g., class labels, points, bounding boxes) to refine the outcome. 

Our {\bf main contributions} are as follows:

\begin{itemize}
    \item {\it New segmentation paradigm:} We propose the new automatic and interactive segmentation paradigm and introduce the first model within this paradigm.
    \item {\it Functionality enhancement:} We devise the first automated point prompt generation module and the corresponding specialized loss functions.
    \item {\it Empirical validation:} We performed comprehensive quantitative and qualitative experiments on various datasets to evaluation the effectiveness of our method.
\end{itemize}

%% file: sec/2_relatedworks.tex
\section{Related Works}
\label{sec:relatedworks}

\textbf{Automatic Semantic Segmentation.} The domain of semantic segmentation has experienced growth and transformation. Earlier work~\cite{long2015fully,ronneberger2015u,yu2018bisenet,fan2021rethinking,badrinarayanan2017segnet,chen2014semantic,chen2017deeplab,chen2017rethinking,chen2018encoder,zhu2021learning,zhao2017pyramid} primarily used convolutional neural networks. Subsequent innovations saw the integration of attention modules~\cite{zhang2019acfnet,fu2019dual,zhu2019asymmetric}. Reflecting a broader trend in computer vision, more recent developments have pivoted towards transformer-based models~\cite{zheng2021rethinking,xie2021segformer,strudel2021segmenter,cheng2022masked,chen2022vision}, often leveraging large-scale pre-trained models~\cite{wang2023one,wang2023internimage,su2023towards,fang2023eva}. These methods are all automated in nature.

\noindent\textbf{Adaptation of Interactive Segmentation Models.} Interactive segmentation models~\cite{liu2023simpleclick,lin2023adaptiveclick,kirillov2023segany} take a prompt from the user and generate the corresponding segmentation map. These models are designed to be generalized to many tasks. Unlike traditional automatic segmentation models, which are limited to predefined classes, interactive models, in theory, can handle any class. SAM~\cite{kirillov2023segany} has emerged as a prominent interactive model due to its exceptional performance across multiple domains. Subsequent research has focused on enhancing SAM's interactive segmentation capability in specific domains through fine-tuning with ground truth prompts and segmentation pairs~\cite{ma2023segment,wu2023medical,dai2023samaug,ke2023segment,deng2023segment}. Additionally, other studies have explored extending SAM to automatic segmentation, through workaround~\cite{lei2023medlsam,liu2023grounding} or through automatic adaptation methods~\cite{zhang2023customized,chen2023sam,cui2023all,paranjape2023adaptivesam,chen2023ma,xiong2023mammo,hu2023efficiently,shaharabany2023autosam}. However, there remains a gap in adapting interactive segmentation models to an automatic segmentation setting while retaining their promptability.


%% file: sec/3_method.tex
\section{Preliminary}
\label{sec:property_prompt}

\input{latex_figs/heatmaps}

Despite the variety of prompts accepted by SAM, the difference among these prompts in terms of their ability to generate good segmentation results is under-explored. Given that the efficacy of interactive segmentation methods is intrinsically linked to prompt quality, it is essential to systematically analyze various prompts. In this section, we aim to establish a theoretical framework to evaluate the mistake behaviors of different prompt types and demonstrate an example application of the framework to SAM. Our analysis will predominantly focus on bounding box and point prompts, as they are the most commonly utilized visual prompts in this context. The goals of this section are to evaluate different prompts and to select a suitable prompt in our setting.

\noindent\textbf{Confusion Matrix as the Framework.} Visual prompts facilitate the model's differentiation of intended and unintended behaviors, functioning analogously to a classifier. For a prompt to be effective, it must clearly separate class semantics. Analogous to the confusion matrix used in classifier evaluation, we introduce the concept of a \textbf{Prompt Confusion Matrix} (PCM). It features True Semantic Similarity (TSS) along its diagonal, representing correct semantic interpretation, and False Semantic Similarity (FSS) in its off-diagonal elements, indicating semantic misinterpretations. Furthermore, the precision of the generated segmentation should be measure in order to compare the effect of prompts on the output. We quantify it using the \textbf{Output Confusion Matrix} (OCM), where True Output Similarity (TOS) is placed on the diagonal, and False Output Similarity (FOS) takes the off-diagonal positions. A functional PCM correlates with the OCM, since we assume the better separation of the semantic the prompts provides, the better segmentation the model generates. It is clear that PCM and OCM function like the conventional confusion matrix where the we aim for high diagonal values but low off-diagonal values. While other metrics derived from the confusion matrix might be simpler for benchmarking, they often overlook certain error behaviors. Thus, our proposed PCM and OCM are more suitable as evaluation tools. Various implementations of PCM and OCM based on different architectures should still adhere to the core concept of the confusion matrix.

\noindent\textbf{Implementation of PCM and OCM.} If a prompt effectively separates image features into distinct groups, differences between these group features should be evident in the PCM.We define the image feature $X_i$ of class $i$ and prompt features $P_j$ of class $j$, where $X_i$ comprises all encoded image patches with location information within the segmentation mask of class $i$, and $P_j$ is an aggregated representation of the prompt. Each point prompt $P_j$ is represented by the nearest image patch, and each bounding box prompt by averaging the image patches within the box; this averaging approach is adopted because SAM represents an entire box as a single prompt. We define semantic similarity $s_{i,j}=\mathrm{mean}_{x_i\in X_i}(\mathrm{sim}(x_i,p_{x_i}^{\max}))$ as the mean of the highest similarities between each image feature $x_i$ and its most similar prompt feature $p_{x_i}^{\max}=\max_{p_j\in P_j}(\mathrm{sim}(x_i,p_j))$, where $\mathrm{sim}$ denotes cosine similarity. Additionally, output similarity is defined as the overlap ratio between the segmentation masks of two classes, closely mirroring the segmentation model's performance. Due to the significant impact of different implementations on results, our implementation should not serve as a metric to quantify prompt effectiveness; it is solely used to reveal the error behaviors of different prompt types. Based on our implementation, we propose the following:
\begin{enumerate}[label=\textit{Proposition \arabic*}:, align=left]
  \item PCM correlates with OCM.
  \item For the same model, adding the same type of prompts cannot decrease TSS/FSS.
\end{enumerate}

We empirically validate Proposition 1 with the ACDC medical image dataset (the dataset is detailed in Sec.~\ref{sec:dataset}). The point prompts are randomly sampled from the ground truth segmentation mask and the box prompts are the tightest bounding box around the ground segmentation mask. From Fig.~\ref{fig:conf_mat}, we observed a clear correlation between PCM and OCM which supports our Proposition 1. Moreover, Proposition 2 is evident since adding points enhances TSS but not FSS as shown by the PCM's transition from one to four points in Fig.~\ref{fig:conf_mat}. Proposition 2 is also proved by contradiction: assuming adding a prompt $\hat{p_j}$ decreases TSS/FSS. To change the TSS/FSS for any $x_i$, $\max_{p_j\in P_j}(\mathrm{sim}(x_i,p_j))=\mathrm{sim}(x_i,\hat{p_j})>\mathrm{sim}(x_i,p_j)$ which lead to an increase in TSS/FSS, thus contradicting the assumption.

Based on Proposition 1, achieving effective segmentation results requires high TSS and low FSS. Proposition 2 suggests that while we cannot reduce FSS by adding prompts, we can enhance TSS. Therefore, point prompts, with their lower FSS, appear to be a more viable prompt type. Additionally, we note that the prompt semantics for class Myo are often similar to those for LV, leading to confusion in the OCM. This issue is more pronounced with box prompts. Interestingly, this finding aligns with SAM's prompt design, where either only point prompts are used or background point prompts are allowed in addition to the box prompt to reduce confusion; however, box prompts are never used in isolation. Based on these insights, our method will exclusively use point prompts for simplicity. 

\section{Method}
\label{sec:method}

We introduce the AI-SAM, which comprises the Automatic and Interactive Prompter (AI-Prompter) for automatic prompt generation, heuristic-based prompt loss functions, a classifier to exclude prompts from non-existing classes, and a SAM-based interactive segmentation model. The AI-Prompter, steered by the prompt loss functions, is trained to generate a set of easily modifiable and correctable point prompts for a given image and target class. We detail the overall architecture of AI-SAM and the properties that guide the AI-Prompter training process in subsequent sections. The details of our custom classifier, designed to leverage encoded image features, are deferred to the Appendix due to its supportive role rather than being the core contribution of this work.

\begin{figure*}[ht]
    \centering
    \includegraphics[width=\textwidth]{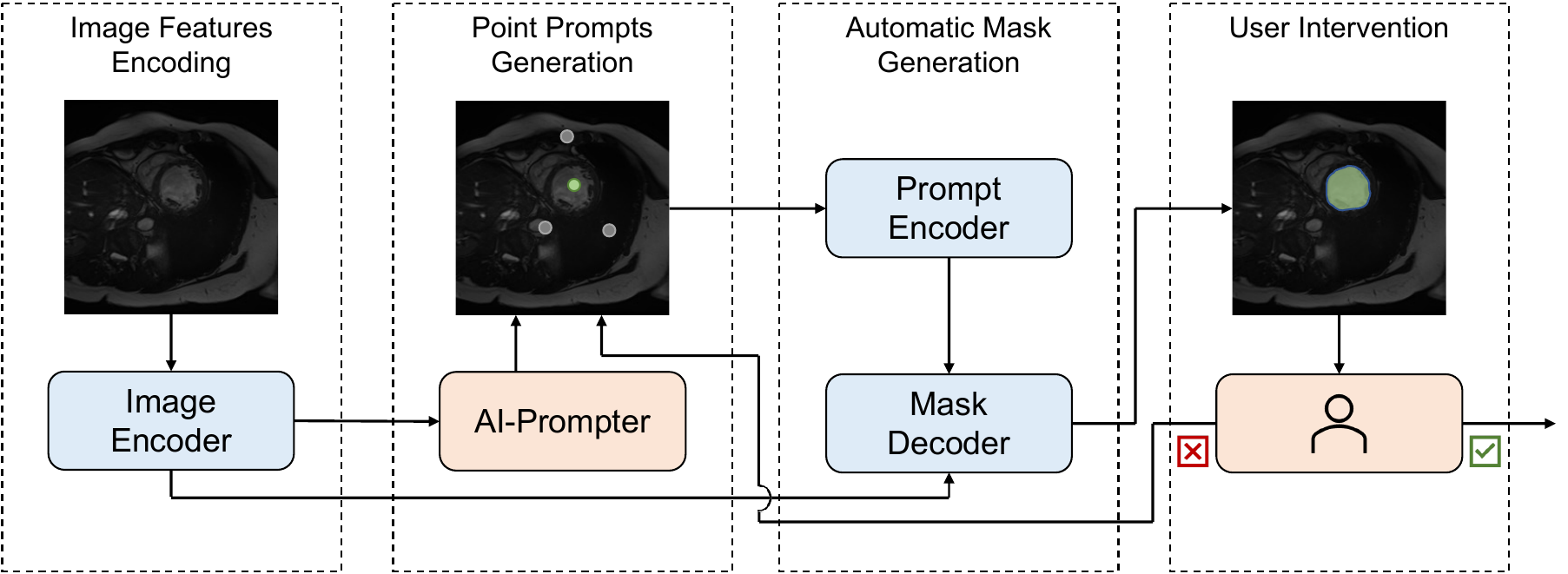}
    \caption{The pipeline of AI-SAM showing the interaction between different modules. Users have the ability to review the segmentation results and adjust prompts accordingly. The green point is the foreground point and the gray points are the background point.}
    \label{fig:ai-sam-architecture}
\end{figure*}

\subsection{Automatic and Interactive Segmentation}

We now summarize both automatic and interactive segmentation methods. Fig.~\ref{fig:sam_overview} is an illustration of the different segmentation methods.  Automatic segmentation models autonomously produce segmentation masks for all predefined classes from an input image. In contrast, interactive segmentation models require both an input image and user-provided prompts to generate the corresponding segmentation mask. The distinction primarily lies in their approach to class semantics: automatic segmentation integrates class semantics into mask generation, whereas interactive segmentation relies solely on the semantics represented by user prompts, regardless of inherent class semantics. Therefore, a model that integrates both automatic and interactive features should bridge class semantics with prompt semantics. The essence of the AI-SAM is its ability to simulate the human prompting process for each predefined class during automatic training, thereby preserving the model's awareness of prompts. The AI-Prompter, crafted to mimic human-like prompt generation, processes image features and generates point prompts for the mask decoder during training. In inference, AI-SAM independently generates prompts and corresponding segmentation masks for predefined classes, while also accommodating user adjustments, such as adding or removing points, to refine segmentation quality. AI-SAM is designed for end-to-end training, utilizing ground truth segmentation masks as targets. Its architecture is depicted in Fig.~\ref{fig:ai-sam-architecture}.

\subsection{Automatic and Interactive Prompter}

Despite point prompts are shown to cause less confusion, as shown in Sec.~\ref{sec:property_prompt}, their potential in interactive adaptation remains largely untapped, possibly due to two main issues. First, there's no clear consensus on what constitutes an optimal point. Unlike bounding boxes, where the ideal is the smallest encompassing an object, there's no analogous standard for points. Prior research in different domains~\cite{toshev2014deeppose,sun2018integral,chen2020ai} has typically regressed points using a Gaussian distribution around the true location, but such ground truth points are not available in our context, requiring a data-driven approach to infer the optimal points. Furthermore, this method lacks contextual richness, making the learning process challenging. To overcome this, we draw inspiration from the pose estimation field's heatmap representation~\cite{tompson2014joint} and introduce the concept of a `generalized point.'

Given a point embedding $p_{i}$ on an image (i.e., the positional embedding), we define generalized point representation as the weighted sum of these positional embeddings, represented as $P=[p_{0},...p_{I}]$ using weights $W=[w_{0},...,w_{I}]$, where $I$ is the number of positional embeddings in a image. That is, $P^\text{g}=W^\top P$.   As the base interactive model provides the representation for positional embeddings, our task is reduced to modeling the generation of $W$. To differentiate from the traditional encoding of points, we refer to the conventional representation as a one-hot point, which is a special case of the generalized point with $W$ being one-hot.

\begin{figure}[ht!]
    \centering
    \includegraphics[width=\linewidth]{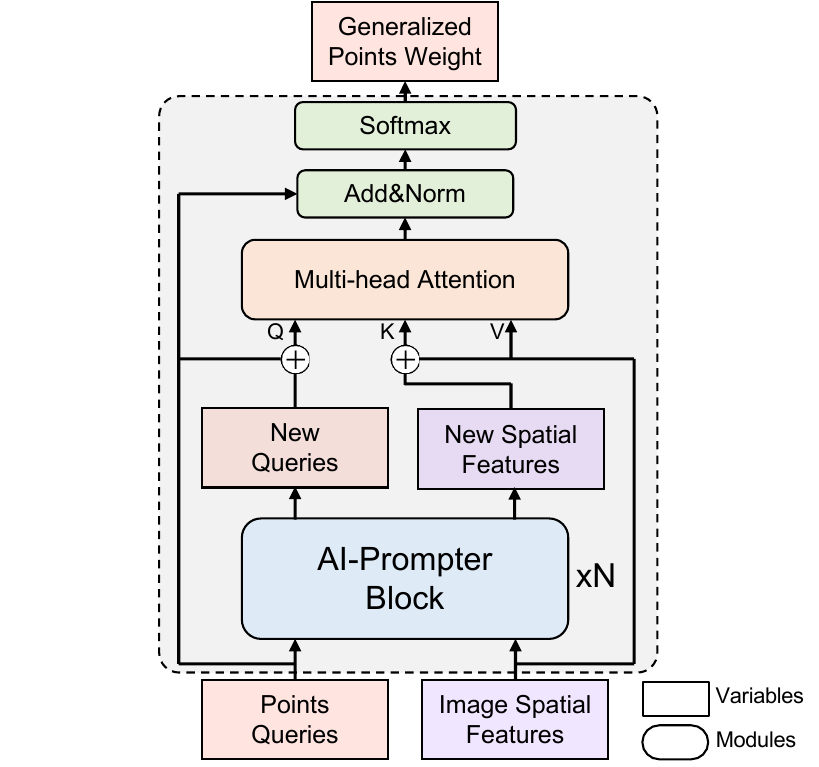}
    \caption{The structure of the AI-Prompter.}
    \label{fig:ai-prompter}
\end{figure}
\begin{figure}[ht!]
    \centering
    \includegraphics[width=\linewidth]{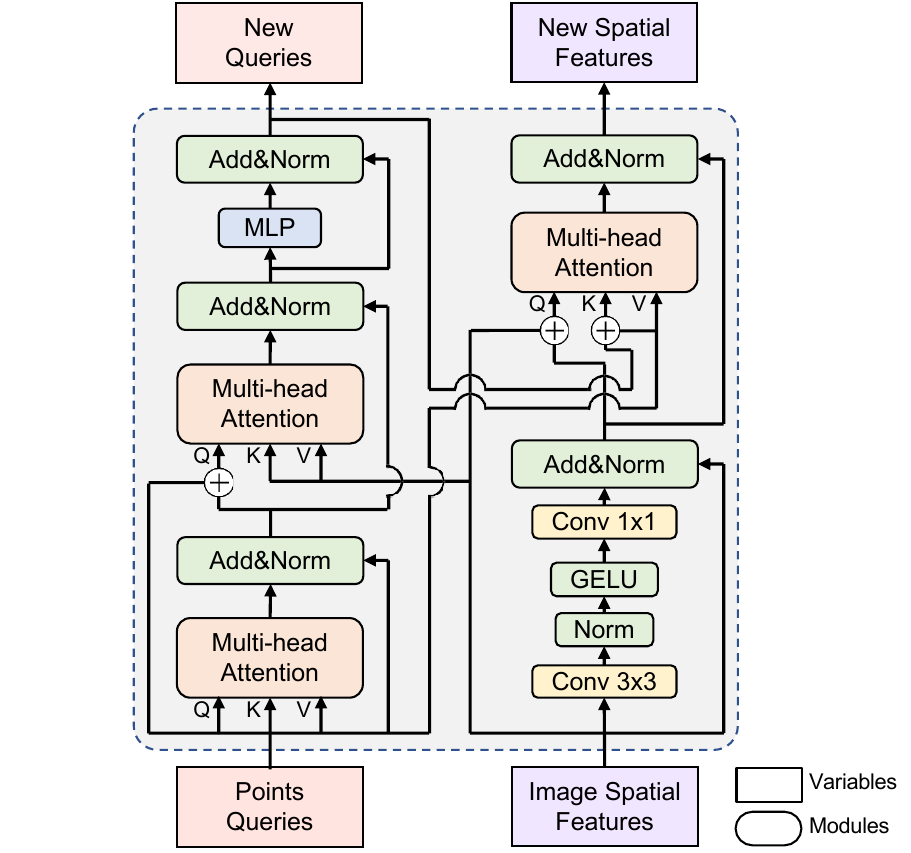}
    \caption{The illustration of one AI-Prompter block.}
    \label{fig:ai-prompter-block}
\end{figure}

We thus propose the AI-Prompter, which processes image feature $X$ and produces weights $W=\mathrm{AIPrompter}(X,c)$, where $c$ denotes the class. The differentiable nature of $P^\text{g}$  allows it to be fed directly into the prompt encoder as $P^\text{g}=\mathrm{AIPrompter}(X,c)^\top P$, facilitating the application of any automatic adaptation method. The entire model is trained end-to-end, using only the ground truth segmentation masks as targets. The AI-Prompter utilizes a customized architecture, the specifics of which are elaborated in Fig.~\ref{fig:ai-prompter}. It resembles SAM's two-way transformer but differs mainly in its two-way attention block design. While SAM employs positional embeddings in the manner of a typical transformer, our approach adopts convolution layers, as illustrated in Fig.~\ref{fig:ai-prompter-block}. This choice is motivated by two key considerations. First, distinguishing between nearby image tokens is more crucial and challenging than further apart. Convolution layers place greater emphasis on discerning points in the vicinity of each other. Second, the relative position of each image patch takes precedence over its absolute position. Convolution layers inherently provide positional cues since they operate only on the neighborhoods of points.

\subsection{Prompt Heuristic Loss}
\label{sec:phl}

As both an automatic and interactive model, AI-SAM is designed with a focus on usability as well as performance. First and foremost, there is no assurance that the generated generalized points will be accurately positioned within the object of interest. Neural networks can take shortcuts, relying on class representation instead of precise point prompts to adapt automatically. If the generated points are linked to class representation without correctly attending to the object's location, usability is entirely compromised. Therefore, ensuring that produced points are correctly associated with the target object location is vital.

Moreover, although the proposed generalized point representation simplifies training, modifying these points can pose challenges due to the disparity between one-hot points and generalized points. Users typically find modifying one-hot points intuitive, as it involves directly clicking on an image. However, modifying a generalized point, which may correspond to multiple parts of the image, can be as complex as altering the segmentation mask itself. Hence, producing generalized points that closely resemble the simplicity and directness of one-hot points is equally crucial.

To address these issues, we introduce two heuristics based on the annotators' intuitions: 
\begin{itemize}
\item $P^\text{g}$ should be situated within the object.
\item $P^\text{g}$ should closely resemble a one-hot point prompt.
\end{itemize}
These heuristics lead to the definition of two loss functions: point correctness loss $\mathcal{L}^\text{pc}=\mathrm{mean}_{c}\mathcal{L}^\text{pc}_c$ and point sharpness loss $\mathcal{L}^\text{ps}=\mathrm{mean}_{c}\mathcal{L}^\text{ps}_c$, which are defined as follows:
\begin{eqnarray}
\mathcal{L}^\text{pc}_c&=&1-\frac{\mathbf{1}_{c}^\top W+\gamma}{\mathbf{1}^\top W+\gamma}\;,\\
\mathcal{L}^\text{ps}_c&=&1-\frac{\mathrm{max}(\mathbf{1}_{c}\odot W)+\gamma}{\mathbf{1}_{c}^\top W+\gamma}\;,
\end{eqnarray}
where $\mathbf{1}_{c}$ is an indicator function with its $i$th element being $1$ if the image feature $x_i$ belongs to class $c$ and $0$ otherwise, $\mathbf{1}$ is a matrix of ones the same size as $W$, $\odot$ represents element-wise multiplication, and $\gamma$ is a hyper-parameter. $\mathcal{L}^\text{pc}_c$ penalizes the difference between the total weight $\mathbf{1}^\top W$ and the correct weight $\mathbf{1}_{c}^\top W$ thus encourage correct $P^\text{g}$. $\mathcal{L}^\text{ps}_c$ penalizes the difference between the correct weight $\mathbf{1}_{c}^\top W$ and the maximum correct weight $\mathrm{max}(\mathbf{1}_{c}\odot W)$ thus encourage one-hot points. The combination of $\mathcal{L}^\text{pc}$ and $\mathcal{L}^\text{ps}$ ensures the final $P^\text{g}$ is both inside the object and closely aligned with a one-hot point. 

Without regularization, the aforementioned loss would cause all points to converge to the same optimal point, making it impossible to control the number of points. As a generalized point can encapsulate information from many locations, multiple points might seem redundant. However, to enhance usability, the generalized points are encouraged to be one-hot, thereby precluding them from representing information from multiple locations. More importantly, representing all prompt information in one generalized point would require users to modify the entire class prompt simultaneously; distributing the information across multiple prompts allows for more fine-grained control. Therefore, we must design a robust heuristic for diverse point prompt generation.

Based on Proposition 1, which states that adding points can improve performance, we aim to encourage the model to produce multiple points with high TSS and low FSS. To achieve these goals, we introduce the prompt diversity loss
\begin{equation}
\mathcal{L}^\text{pd}=\beta^\text{in}\mathcal{L}^\text{in}+\beta^\text{out}\mathcal{L}^\text{out}\;,
\end{equation}
consisting of the inter-class diversity loss $\mathcal{L}^\text{in}=\mathrm{mean}_{c}\mathcal{L}^\text{in}_c$ and intra-class diversity loss $\mathcal{L}^\text{out}=\mathrm{mean}_{n}\mathcal{L}^\text{out}_n$.  We utilize $\hat{P}^\text{g}=W^{\top}(P+X)$ as the point feature, considering both image features and positional embeddings as important for distinguishing optimal points. These loss components are defined as follows:
\begin{eqnarray}
\mathcal{L}^\text{in}_c&=&-\log\frac{\exp(1/\tau)}{\sum_n \exp(\mathrm{sim}(\hat{p}_{n^+}^\text{g},w_{n})/\tau)}\;,\\
\mathcal{L}^\text{out}_n&=&-\log\frac{\exp(1/\tau)}{\sum_c \exp(\mathrm{sim}(\hat{p}_{c^+}^\text{g},w_{c})/\tau)}\;,
\end{eqnarray}
where $c^+$ and $n^+$ denote the class index and prompt index of the anchor point, respectively.
$\mathcal{L}^\text{in}$ penalizes similarity among points within the same class, contributing to TSS, as similar points do not improve TSS.  Conversely, $\mathcal{L}^\text{in}$ penalizes similarity between points of different classes, contributing to FSS, as incorrect points can significantly increase FSS.  The final prompt heuristic loss is defined as: 
\begin{equation}
    \mathcal{L}^{ph}=\alpha^\text{pc}\mathcal{L}^\text{pc}+\alpha^\text{ps}\mathcal{L}^\text{ps}+\alpha^\text{pd}\mathcal{L}^\text{pd},
\end{equation}
where $\alpha^\text{pc}$, $\alpha^\text{ps}$, and $\alpha^\text{pd}$ represent the weights assigned to each respective loss components.

%% file: latex_figs/heatmaps.tex
\begin{figure}[ht!]
\centering 

\begin{tikzpicture}[scale=0.8]
\begin{axis}[
    width=0.54\linewidth,
    height=0.54\linewidth,
    view={0}{90},
    xlabel=1 Point,
    ylabel=Prompt Feature,
    colormap={bluewhite}{color=(white) rgb255=(90,96,191)}, 
    xtick={0,...,3},
    ytick={0,...,3},
    xticklabels={None, RV, Myo, LV},
    yticklabels={None, RV, Myo, LV},
    enlargelimits=false,
    axis on top,
    xticklabel style={anchor=north, rotate=0},
    yticklabel style={anchor=south, rotate=90},
    axis line style={draw=none},tick style={draw=none}, 
    mesh/cols=4, 
    mesh/rows=4, 
    point meta min=0, 
    point meta max=1, 
    nodes near coords align={anchor=center},
    nodes near coords={\pgfmathprintnumber[/pgf/number format/fixed]\pgfplotspointmeta},
]

\addplot[matrix plot*, point meta=explicit] table [meta=C] {
    x y C
    0 0 0.1861
    1 0 0.323
    2 0 0.1953
    3 0 0.1942
    0 1 0.1673
    1 1 0.7473
    2 1 0.366
    3 1 0.364
    0 2 0.168
    1 2 0.4825
    2 2 0.5408
    3 2 0.5925
    0 3 0.1625
    1 3 0.4709
    2 3 0.5874
    3 3 0.7385
};
\end{axis}

\begin{axis}[
    at={(0.37\linewidth,0)}, 
    width=0.54\linewidth,
    height=0.54\linewidth,
    view={0}{90},
    xlabel=4 Points,
    colormap={bluewhite}{color=(white) rgb255=(90,96,191)}, 
    xtick={0,...,3},
    xtick={0,...,3},
    ytick={0,...,3},
    xticklabels={None, RV, Myo, LV},
    yticklabels={},
    enlargelimits=false,
    axis on top,
    tick label style={font=\small},
    xticklabel style={anchor=north, rotate=0},
    yticklabel style={anchor=south, rotate=90},
    axis line style={draw=none},tick style={draw=none}, 
    mesh/cols=4, 
    mesh/rows=4, 
    point meta min=0, 
    point meta max=1, 
    nodes near coords align={anchor=center},
    nodes near coords={\pgfmathprintnumber[/pgf/number format/fixed]\pgfplotspointmeta},
]

\addplot[matrix plot*, point meta=explicit] table [meta=C] {
    x y C
    0 0 0.2656
    1 0 0.3693
    2 0 0.2504
    3 0 0.2379
    0 1 0.1734
    1 1 0.9052
    2 1 0.451
    3 1 0.4411
    0 2 0.1923
    1 2 0.6701
    2 2 0.8018
    3 2 0.7784
    0 3 0.1708
    1 3 0.5847
    2 3 0.7747
    3 3 0.8958
};

\end{axis}
\begin{axis}[
    at={(0.74\linewidth,0)}, 
    width=0.54\linewidth,
    height=0.54\linewidth,
    view={0}{90},
    xlabel=Bounding Box,
    colormap={bluewhite}{color=(white) rgb255=(90,96,191)}, 
    xtick={0,...,3},
    ytick={0,...,3},
    xticklabels={None, RV, Myo, LV},
    yticklabels={},
    enlargelimits=false,
    axis on top,
    tick label style={font=\small},
    xticklabel style={anchor=north, rotate=0},
    yticklabel style={anchor=south, rotate=90},
    axis line style={draw=none},tick style={draw=none}, 
    mesh/cols=4, 
    mesh/rows=4, 
    point meta min=0, 
    point meta max=1, 
    nodes near coords align={anchor=center},
    nodes near coords={\pgfmathprintnumber[/pgf/number format/fixed]\pgfplotspointmeta},
]

\addplot[matrix plot*, point meta=explicit] table [meta=C] {
    x y C
    0 0 0.3562
    1 0 0.4594
    2 0 0.3467
    3 0 0.3407
    0 1 0.1775
    1 1 0.8652
    2 1 0.4718
    3 1 0.5028
    0 2 0.1817
    1 2 0.5824
    2 2 0.7314
    3 2 0.8604
    0 3 0.1723
    1 3 0.5457
    2 3 0.6999
    3 3 0.8827
};

\end{axis}
\begin{axis}[
    at={(0,0.37\linewidth)}, 
    width=0.54\linewidth,
    height=0.54\linewidth,
    view={0}{90},
    ylabel=Output Feature,
    colormap={bluewhite}{color=(white) rgb255=(90,96,191)}, 
    xtick={0,...,3},
    ytick={0,...,3},
    xticklabels={},
    yticklabels={None, RV, Myo, LV},
    enlargelimits=false,
    axis on top,
    xticklabel style={anchor=north, rotate=0},
    yticklabel style={anchor=south, rotate=90},
    axis line style={draw=none},tick style={draw=none}, 
    mesh/cols=4, 
    mesh/rows=4, 
    point meta min=0, 
    point meta max=1, 
    nodes near coords align={anchor=center},
    nodes near coords={\pgfmathprintnumber[/pgf/number format/fixed]\pgfplotspointmeta},
]

\addplot[matrix plot*, point meta=explicit] table [meta=C] {
    x y C
    0 0 0.1137
    1 0 0.1798
    2 0 0.0365
    3 0 0.0404
    0 1 0.0105
    1 1 0.8595
    2 1 0.1494
    3 1 0.1592
    0 2 0.0114
    1 2 0.2639
    2 2 0.4879
    3 2 0.6801
    0 3 0.0069
    1 3 0.1908
    2 3 0.2658
    3 3 0.9038
};
\end{axis}

\begin{axis}[
    at={(0.37\linewidth,0.37\linewidth)}, 
    width=0.54\linewidth,
    height=0.54\linewidth,
    view={0}{90},
    colormap={bluewhite}{color=(white) rgb255=(90,96,191)}, 
    xtick={0,...,3},
    ytick={0,...,3},
    xticklabels={},
    yticklabels={},
    enlargelimits=false,
    axis on top,
    tick label style={font=\small},
    xticklabel style={anchor=north, rotate=0},
    yticklabel style={anchor=south, rotate=90},
    axis line style={draw=none},tick style={draw=none}, 
    mesh/cols=4, 
    mesh/rows=4, 
    point meta min=0, 
    point meta max=1, 
    nodes near coords align={anchor=center},
    nodes near coords={\pgfmathprintnumber[/pgf/number format/fixed]\pgfplotspointmeta},
]

\addplot[matrix plot*, point meta=explicit] table [meta=C] {
    x y C
    0 0 0.2737
    1 0 0.1791
    2 0 0.0377
    3 0 0.0379
    0 1 0.0066
    1 1 0.8856
    2 1 0.0807
    3 1 0.0594
    0 2 0.0073
    1 2 0.2352
    2 2 0.5114
    3 2 0.4471
    0 3 0.0036
    1 3 0.188
    2 3 0.2553
    3 3 0.9365
};

\end{axis}
\begin{axis}[
    at={(0.74\linewidth,0.37\linewidth)}, 
    width=0.54\linewidth,
    height=0.54\linewidth,
    view={0}{90},
    colormap={bluewhite}{color=(white) rgb255=(90,96,191)}, 
    xtick={0,...,3},
    ytick={0,...,3},
    xticklabels={},
    yticklabels={},
    enlargelimits=false,
    axis on top,
    tick label style={font=\small},
    xticklabel style={anchor=north, rotate=0},
    yticklabel style={anchor=south, rotate=90},
    axis line style={draw=none},tick style={draw=none}, 
    mesh/cols=4, 
    mesh/rows=4, 
    point meta min=0, 
    point meta max=1, 
    nodes near coords align={anchor=center},
    nodes near coords={\pgfmathprintnumber[/pgf/number format/fixed]\pgfplotspointmeta},
]

\addplot[matrix plot*, point meta=explicit] table [meta=C] {
    x y C
    0 0 0.4964
    1 0 0.4293
    2 0 0.3178
    3 0 0.3336
    0 1 0.0013
    1 1 0.9628
    2 1 0.0879
    3 1 0.0447
    0 2 0.0012
    1 2 0.264
    2 2 0.8777
    3 2 0.9985
    0 3 0.0
    1 3 0.1745
    2 3 0.1106
    3 3 0.9224
};
\end{axis}

\end{tikzpicture}

\caption{The Output Confusion Metrices (Top Row) and the Prompt Confusion Metrices (Bottom Row) using SAM feature on ACDC dataset. Values are normalized to a 0-1 scale and averaged over the entire dataset. The class names for individual organ are detailed in Sec.~\ref{sec:dataset}.}
\label{fig:conf_mat}
\end{figure}

%% file: sec/4_results.tex
\section{Experiments}

To validate our approach, we conducted extensive experiments using multiple public-domain datasets and compared with other SOTA models. We report the results in this section. We will also later discuss the potential use of our approach in other real-world medical contexts.

\subsection{Datasets}
\label{sec:dataset}

\noindent\textbf{Synapse Multi-organ dataset\footnote{https://www.synapse.org/\#!Synapse:syn3193805/wiki/217789/}} contains 30 abdominal CT scans. We follow~\cite{chen2021transunet,rahman2023medical} to use the same 12 scans for testing and the rest for training and segmentation labels for 8 abdominal organs: aorta, gallbladder (GB), left kidney (KL), right kidney (KR), liver, pancreas (PC), spleen (SP), and stomach (SM).

\noindent\textbf{Automated Cardiac Diagnosis Challenge (ACDC)\footnote{https://www.creatis.insa-lyon.fr/Challenge/acdc/}} dataset contains 100 cardiac MRI scans each labeled with left ventricle (LV), right ventricle (RV) and myocardium (Myo). We follow~\cite{chen2021transunet,rahman2023medical} to use the same 20 scans for testing but the rest for training.

\noindent\textbf{Camouflaged Object Detection} contains three dataset of RGB images: COD10K~\cite{fan2020camouflaged}, CHAMELEON~\cite{skurowski2018animal}, and CAMO~\cite{le2019anabranch} datasets. COD10K contains 3,040 training and 2,026 testing samples; CAMO contains 1,000 training and 250 testing samples; CHAMELEON contains 76 testing samples. All the samples are labeled with foreground masks. We follow~\cite{fan2020camouflaged} to use the 4,040 training samples from both COD10K and CAMO for training and test on the testing samples for the three datasets

\noindent\textbf{Image Shadow Triplets Dataset (ISTD)}~\cite{wang2018stacked} contains 1,330 training samples and 540 testing samples of RGB images with foreground masks.

The prepossessing steps and evaluation metrics for each dataset follow the corresponding previous work and are detailed in the Appendix.

\begin{table*}[h!]
\centering
\begin{subtable}[t]{1.\textwidth}
{\begin{adjustbox}{width=\textwidth}
\setlength{\tabcolsep}{5pt}
\begin{tabular}{l@{\hskip 3.5pt}l|cccccccccc|cccc}
\toprule
& \multirow{2}{*}{Method} & \multicolumn{10}{c|}{Synapse} & \multicolumn{4}{c}{ACDC} \\
& & DICE $\uparrow$ & HD95$\downarrow$ & Aorta & GB & KL & KR & Liver & PC & SP & SM & DICE $\uparrow$ & RV & Myo & LV \\
\midrule
\parbox[t]{2.3mm}{\multirow{15}{*}{\rotatebox[origin=c]{90}{\textit{Automatic}}}} &
R50+UNet \cite{chen2021transunet}   & 74.68     & 36.87     & 84.18     & 62.84  & 79.19       & 71.29       & 93.35     & 48.23  & 84.41  & 73.92   & 87.55         & 87.10   & 80.63    & 94.92  \\
& R50+AttnUNet \cite{chen2021transunet}    & 75.57     & 36.97     & 55.92     & 63.91  & 79.20       & 72.71       & 93.56     & 49.37  & 87.19  & 74.95  & 86.75         & 87.58   & 79.20    & 93.47  \\
%
& ViT+CUP \cite{chen2021transunet}   & -       & -  & -    & -  & -       & -  & -    & - & -       & -  & 81.45         & 81.46   & 70.71    & 92.18  \\
& R50+ViT+CUP \cite{chen2021transunet}  & -       & -  & -    & -  & -       & -  & -    & - & -       & -  & 87.57         & 86.07   & 81.88    & 94.75   \\
& TransUNet \cite{chen2021transunet}     & 77.48     & 31.69      & 87.23     &  63.13  & 81.87      & 77.02       & 94.08    & 55.86  & 85.08  & 75.62  & 89.71         & 88.86   &  84.53    & 95.73      \\ %
& SwinUNet \cite{cao2022swin}       & 79.13     & 21.55     & 85.47     & 66.53 &  83.28       & 79.61       & 94.29     & 56.58  & 90.66  & 76.60    & 90.00         & 88.55   & 85.62    & 95.83   \\ 
& MT-UNet \cite{wang2022mixed}       & 78.59     & 26.59     & 87.92     & 64.99  & 81.47       & 77.29       & 93.06    & 59.46 & 87.75  & 76.81    & 90.43         & 86.64   & 89.04    & 95.62  \\
& MISSFormer \cite{huang2021missformer}       & 81.96     & 18.20     & 86.99     &  68.65  &  85.21       & 82.00       & 94.41     & 65.67  & 91.92  & 80.81  & 90.86         & 89.55   & 88.04    & 94.99  \\        
& CASTformer \cite{you2022class}       & 82.55     & 22.73     & \textbf{89.05}     &  67.48  & 86.05      & 82.17       & \textbf{95.61}     & 67.49  & 91.00  & 81.55    & -       & -  & -    & - \\
& PVT-CASCADE \cite{rahman2023medical}   & 81.06     & 20.23    & 83.01     & 70.59  & 82.23       & 80.37       & 94.08     & 64.43  & 90.10   & 83.69   & 91.46         & 88.9    & 89.97    & 95.50    \\ %
& TransCASCADE \cite{rahman2023medical}  & 82.68     & 17.34     & 86.63     & 68.48  & 87.66       & 84.56       & 94.43     & 65.33  & 90.79  & 83.52   & 91.63         & 89.14  & \textbf{90.25}    & 95.50 \\
& Parallel MERIT~\cite{rahman2023multi}    & 84.22     & 16.51     & 88.38     & 73.48 & 87.21       & 84.31       & 95.06     & 69.97  & 91.21   & 84.15     & \textbf{92.32}         & \textbf{90.87}    & 90.00    & \textbf{96.08} \\
& Cascaded MERIT~\cite{rahman2023multi}    & \textbf{84.90}     & {13.22}     & 87.71     & {74.40}  & \textbf{87.79}      & {84.85}       & 95.26     & {71.81}  &  \textbf{92.01}   & {85.38}   & 91.85       & 90.23  & 89.53    & 95.80  \\
& SAMed\_h~\cite{zhang2023customized}  & 84.30    & 16.02     & 87.81     & \textbf{74.72}              & 85.76    & 81.52      & 95.76     & 70.63  &  90.46   & \textbf{87.77}   & -       & -  & -    & -  \\
& \textbf{AI-SAM}     & 84.21     & \textbf{12.11}    & 88.89     & 74.53  & 86.56      & \textbf{85.01}       & 96.30     & \textbf{72.84} &  90.32   & 79.24   & 92.06       & 90.18  & 89.94    & 96.05 \\
\midrule
\parbox[t]{2.3mm}{\multirow{5}{*}{\rotatebox[origin=c]{90}{\textit{Interactive}}}} 
 & SAM$^*$ gt box  & 90.16   & 3.27              & 92.47    & \textbf{90.82}     & 91.40             & 90.71    & 92.44     & 76.87     & 95.88 &  \textbf{90.70}   & 79.56      & 88.15               & 57.05  & 93.49  \\
& MedSAM$^*$ gt box  & 88.82   & 2.41              & 90.15    & 82.97     & 90.95              & 89.76    & 95.74      & 76.33     & 94.55 &  90.14  & 67.95      & 92.22               & 16.50   & 95.14  \\
& \textbf{AI-SAM} gt label     & 87.56     & 10.14     & 91.51     & 83.78  & 89.48      & 87.34       & 96.51    & 76.96 &  94.62   & 80.28   & 93.02       & 92.58  & 90.21  & {96.26} \\
& \textbf{AI-SAM} 1 rd pt   & 87.91   & 6.78    & 90.26   & 83.70            & 89.85   & 88.49      & 96.52     & 77.16  &  95.32  & 81.97 & 93.04       & 92.58               & 90.26   & 96.28 \\
& \textbf{AI-SAM} gt box     & \textbf{90.66}   & \textbf{1.73}    & \textbf{95.03}    & 85.20    & \textbf{93.40}  & \textbf{92.13}     & \textbf{96.76}      & \textbf{81.79}     & \textbf{96.27}   &  84.73     & \textbf{93.89}       & \textbf{94.13}  & \textbf{90.95}    & \textbf{96.59}  \\
\bottomrule
\end{tabular}
\end{adjustbox}}
\caption{Medical Image Segmentation.}
\label{tab:medical_sotas}
\end{subtable}
\begin{subtable}[t]{0.765\textwidth}
{\begin{adjustbox}{width=\textwidth}
\centering
\setlength{\tabcolsep}{3.3pt}
\begin{tabular}{l@{\hskip 5pt}l|cccc|cccc|cccc}
\toprule
& \multirow{2}{*}{Method} & \multicolumn{4}{c|}{CHAMELEON  \cite{skurowski2018animal}}   & \multicolumn{4}{c|}{CAMO \cite{le2019anabranch}}        & \multicolumn{4}{c}{COD10K \cite{fan2020camouflaged}}       \\ 
& & $ S_\alpha \uparrow$               & $E_\phi \uparrow$              & $F^\omega_\beta \uparrow$              & MAE $\downarrow$           & $S_\alpha \uparrow$              & $E_\phi \uparrow$              & $F^\omega_\beta \uparrow$               & MAE $\downarrow$           & $S_\alpha \uparrow$              & $E_\phi \uparrow$              & $F^\omega_\beta \uparrow$               & MAE $\downarrow$           \\ \midrule
\parbox[t]{2.3mm}{\multirow{7}{*}{\rotatebox[origin=c]{90}{\textit{Automatic}}}}
& SINet\cite{fan2020camouflaged}    & 0.869          & 0.891          & 0.740          & 0.440          & 0.751          & 0.771          & 0.606          & 0.100          & 0.771          & 0.806          & 0.551          & 0.051          \\
& JCOD \cite{li2021uncertainty}    & 0.870          & 0.924          & -              & 0.039          & 0.792          & 0.839          & -              & 0.82           & 0.800          & 0.872          & -              & 0.041          \\
& PFNet \cite{mei2021camouflaged}   & 0.882          & \textbf{0.942} & 0.810          & 0.330          & 0.782          & 0.852          & 0.695          & 0.085          & 0.800          & 0.868          & 0.660          & 0.040          \\
& FBNet  \cite{lin2023frequency}  & 0.888          & 0.939          & \textbf{0.828} & \textbf{0.032} & 0.783          & 0.839          & 0.702          & 0.081          & 0.809          & 0.889          & 0.684          & 0.035          \\
& SAM~\cite{kirillov2023segany}    & 0.727 &  0.734   &  0.639    &  0.081  & 0.684 & 0.687 & 0.606 & 0.132  &  0.783    &    0.798     & 0.701 & 0.050   \\
& SAM-Adapter~\cite{chen2023sam}     & 0.896 & 0.919          & 0.824          & 0.033          & 0.847 & 0.873          & \textbf{0.765}          & 0.070          & {0.883} & {0.918} & {0.801} & \textbf{0.025} \\ 
& \textbf{AI-SAM}     & \textbf{0.901 } & 0.923          & 0.820          & \textbf{0.032}          & \textbf{0.849} & \textbf{0.880}          & 0.759          & \textbf{0.068}          & \textbf{0.885} & \textbf{0.922} & \textbf{0.806} & \textbf{0.025} \\ 
\midrule
\parbox[t]{2.3mm}{\multirow{2}{*}{\rotatebox[origin=c]{90}{\textit{Inter.}}}}
& {SAM}$^*$ gt box  & {0.829 } & 0.859          & 0.718          & {0.088}          & {0.860} & {0.886}          & 0.797          & {0.074}          & {0.860} & {0.896} & {0.764} & {0.053} \\ 
& \textbf{AI-SAM} gt box     & \textbf{0.914 } & \textbf{0.940}         & \textbf{0.863}          & \textbf{0.026}          & \textbf{0.875} & \textbf{0.900}          & \textbf{0.829}          & \textbf{0.054}          & \textbf{0.900} & \textbf{0.936} & \textbf{0.852} & \textbf{0.020} \\ \bottomrule
\end{tabular}
\end{adjustbox}}
\caption{Camouflage Detection.}
\label{tab:cod_sotas}
\end{subtable}
\begin{subtable}[t]{0.229\textwidth}
    {\begin{adjustbox}{width=\textwidth}
    \centering
    \setlength{\tabcolsep}{3pt}
    \begin{tabular}{l@{\hskip 5pt}l| c }
    \toprule
    
    & \multirow{2}{*}{Method} & ISTD~\cite{wang2018stacked} \\
    & &BER $\downarrow$ \\
    \midrule
    \parbox[t]{2.3mm}{\multirow{7}{*}{\rotatebox[origin=c]{90}{\textit{Automatic}}}}
    & BDRAR \cite{zhu2018bidirectional} & 2.69 \\
    & DSC \cite{hu2018direction} & 3.42\\
    & DSD \cite{zheng2019distraction} & 2.17 \\
    & FDRNet \cite{zhu2021mitigating} & 1.55 \\
    & SAM \cite{kirillov2023segany} & 40.51\\
    & SAM-Adapter~\cite{chen2023sam} & {1.43}\\
    & \textbf{AI-SAM}  & \textbf{1.41}\\
    \midrule
    \parbox[t]{2.3mm}{\multirow{2}{*}{\rotatebox[origin=c]{90}{\textit{Inter.}}}}
    & SAM$^*$ gt box & {13.56}\\
    & \textbf{AI-SAM} gt box  & \textbf{1.17}\\
     \bottomrule
    \end{tabular}
    \end{adjustbox}}
\caption{Shadow Detection.}
\label{tab:shadow_sotas}
\end{subtable}
\caption{Comparison to the SOTA methods on multiple datasets. Results of previous work produced by us are noted with an $^*$. All the metrics follow the previous works and are detailed in Appendix~\ref{sec:metric}. For Medical image segmentation, the class names are detailed in Sec.~\ref{sec:dataset}. $\uparrow$: higher is better, $\downarrow$: lower is better. The best results are in bold. gt box: the tightest bounding box from the ground truth segmentation. 1 rd pt: 1 randomly sampled point from the ground truth segmentation.}
\label{tab:seg_sotas}
\end{table*}

\subsection{AI-SAM in Automatic Image Segmentation}
In this section, we evaluate the model's performance in various automatic semantic segmentation tasks. We also integrate our AI-Prompter with other SAM-based automatic adaptation methods to demonstrate that AI-Prompter can enhance any automatic adaptation method without compromising performance. Additionally, we assess the model's interactive segmentation capabilities by providing additional guidance during inference. Furthermore, we conduct an ablation study to validate the assumptions made in the Method section regarding the AI-Prompter. Finally, we present qualitative results to showcase the effectiveness of both the AI-Prompter and AI-SAM. Due to space constraints, we provide all model implementation details in the Appendix.

\subsubsection{AI-SAM in Medical Image Segmentation}
This section focuses on a significant challenge in medical image analysis, segmentation, within the context of both automatic and interactive settings. The objective of these experiments is to validate the effectiveness of AI-SAM in comparison to SOTA models, evaluating its performance in both automatic and interactive segmentation scenarios. In the automatic setting, only the image is used as input. The interactive setting incorporates additional synthesized user inputs, which are generated from the ground truth segmentation map, to guide the model.

\noindent\textbf{Automatic Medical Image Segmentation.} As shown in Table~\ref{tab:seg_sotas}, in the automatic setting, our method achieves SOTA performance on both the Synapse and ACDC datasets. Additionally, we include another SAM-based adaptation method, SAM\_d, which performs comparably to our approach despite its lack of promptability. These results underscore the effectiveness of the automatic and interactive adaptation paradigm, even when only the automatic function is utilized.

\noindent\textbf{Interactive Medical Image Segmentation.} Furthermore, our method's performance demonstrates significant improvement when used interactively. Initially, providing class labels for each image results in a notable performance increase, as shown in Table~\ref{tab:seg_sotas}. This outcome suggests that a substantial portion of errors originates from classifier misclassifications, potentially underestimating the performance of AI-Prompter in generating accurate point prompts in the automatic setting. Additionally, users can employ AI-SAM similarly to the original SAM, where both points and bounding boxes can be provided. To estimate the impact of additional point prompts, we randomly sample points from the ground truth mask and include them in the prompts. Although the improvement is modest, it is expected to be more substantial in real-world applications where human-provided points are likely to be more informative than random points; additional points are particularly effective when they correct errors in the segmentation map. Regarding user-provided bounding boxes, we use the tightest bounding box around the segmentation mask. Given that the best bounding box is consistent across all models, we can compare the performance of MedSAM (a current SOTA interactive medical image segmentation model), SAM, and our AI-SAM. Our AI-SAM handles bounding box prompts differently, constraining the learned weights $W$ from the AI-Prompter to ensure that generated point prompts fall within the provided box. As shown in the interactive part of Table~\ref{tab:seg_sotas}, our method achieves SOTA performance in interactive evaluation, despite MedSAM having access to a significantly larger training dataset.

\subsubsection{AI-Prompter on Any Automatic Method}
In addition to medical imaging, numerous fields can benefit from the automatic and interactive adaptation method. Therefore, assessing the generalizability of AI-Prompter is of importance. In theory, AI-Prompter can be integrated with any SAM-based automatic adaptation method to introduce promptability. In this experiment, we chose to apply AI-Prompter to SAM-Adapter, which represents the current SOTA SAM-based adaptation method in specific domains with limited data. This aligns with our hypothetical use case for AI-SAM. By inheriting the hyperparameters of AI-Prompter from the medical imaging experiments and keeping the original hyperparameters for SAM-Adapter, we achieved a new SOTA performance in both camouflage segmentation and shadow segmentation tasks. While our primary goal was to introduce promptability to SAM-Adapter, we also observed performance gains beyond this. The exact reasons for these performance gains are beyond the scope of this work, but we hypothesize that the additional prompts align better with the SAM training process, leading to improved adaptation.

Next, we evaluated the interactive performance on both Camouflage Object Detection (COD) and shadow detection. Because each image contains only one class, there is no need for a classifier or ground truth labels as prompts. Furthermore, we already know that adding random points to the correct mask does not significantly improve performance. Therefore, we directly assessed the performance of AI-SAM using the tightest bounding box. As shown in the Interactive experiments in Table~\ref{tab:seg_sotas}, incorporating the ground truth mask as a prompt led to a substantial improvement in the performance of AI-SAM. However, SAM did not perform as well; it underperformed the automatic methods despite having access to more information. These results further underscore the importance of an automatic and interactive adaptation method.

\subsubsection{Qualitative Evaluation of AI-SAM}

\begin{figure*}[t!]
    \centering
     \begin{subfigure}[b]{0.195\linewidth}
         \centering
         \includegraphics[width=\linewidth, trim=100pt 100pt 60pt 60pt, clip]{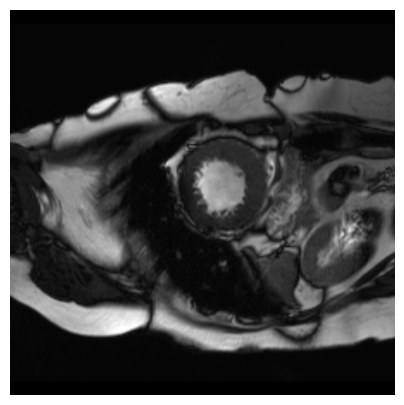}
         \includegraphics[width=\linewidth, trim=100pt 100pt 60pt 60pt, clip]{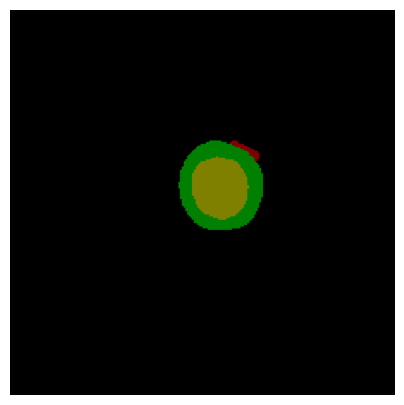}
         \includegraphics[width=\linewidth, trim=40pt 40pt 40pt 40pt, clip]{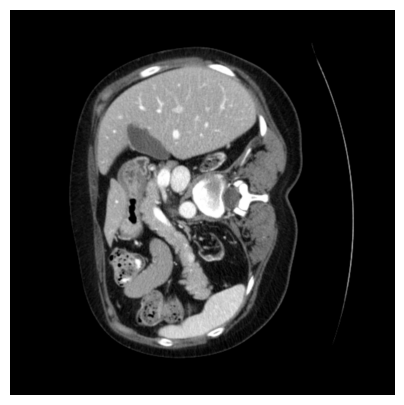}
         \includegraphics[width=\linewidth, trim=40pt 40pt 40pt 40pt, clip]{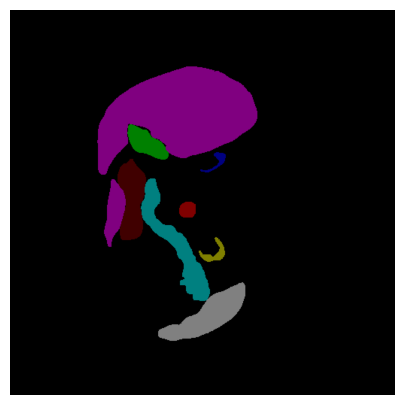}
         \caption{Ground Truth}
     \end{subfigure}
     \begin{subfigure}[b]{0.195\linewidth}
         \centering
         \includegraphics[width=\linewidth, trim=100pt 100pt 60pt 60pt, clip]{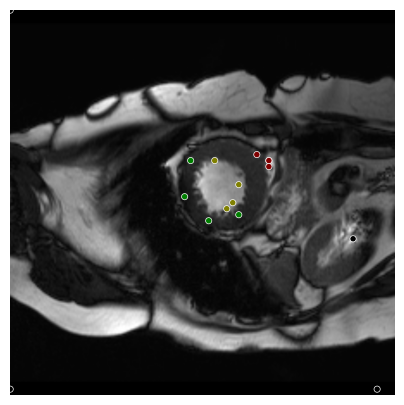}
         \includegraphics[width=\linewidth, trim=100pt 100pt 60pt 60pt, clip]{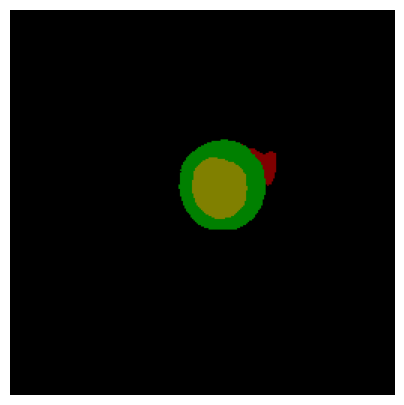}
         \includegraphics[width=\linewidth, trim=40pt 40pt 40pt 40pt, clip]{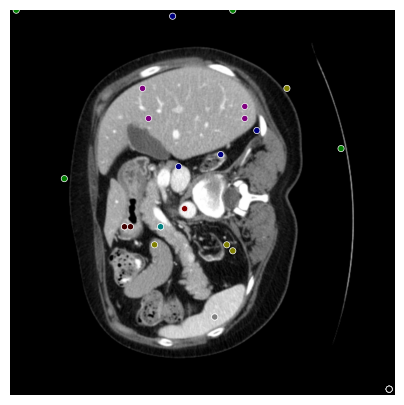}
         \includegraphics[width=\linewidth, trim=40pt 40pt 40pt 40pt, clip]{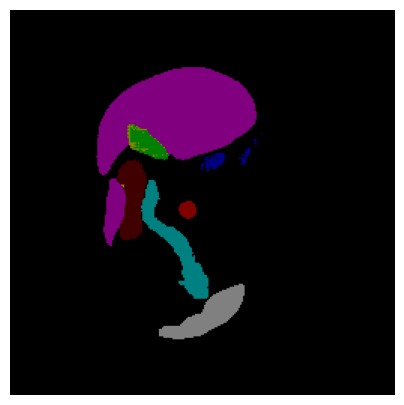}
         \caption{AI-SAM Automatic}
     \end{subfigure}
     \begin{subfigure}[b]{0.195\linewidth}
         \centering
         \includegraphics[width=\linewidth, trim=100pt 100pt 60pt 60pt, clip]{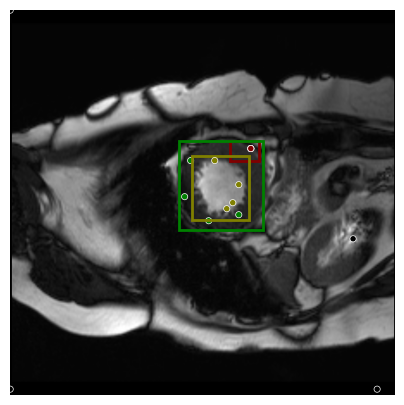}
         \includegraphics[width=\linewidth, trim=100pt 100pt 60pt 60pt, clip]{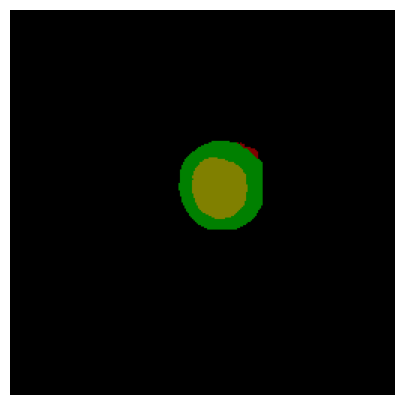}
         \includegraphics[width=\linewidth, trim=40pt 40pt 40pt 40pt, clip]{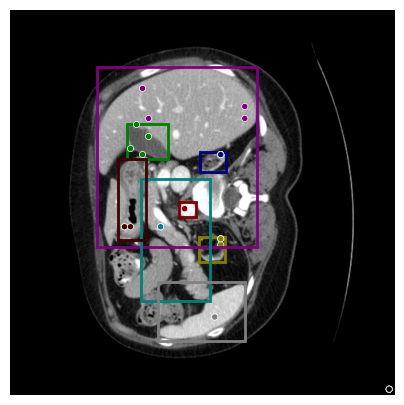}
         \includegraphics[width=\linewidth, trim=40pt 40pt 40pt 40pt, clip]{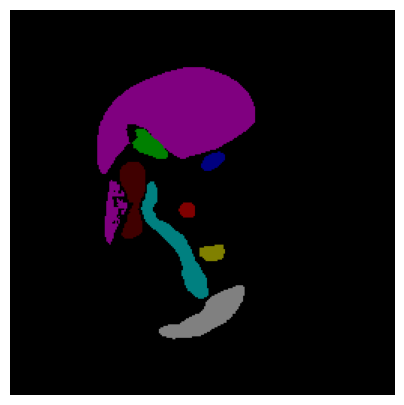}
         \caption{AI-SAM GT Box}
     \end{subfigure}
    \begin{subfigure}[b]{0.195\linewidth}
         \centering
         \includegraphics[width=\linewidth, trim=100pt 100pt 60pt 60pt, clip]{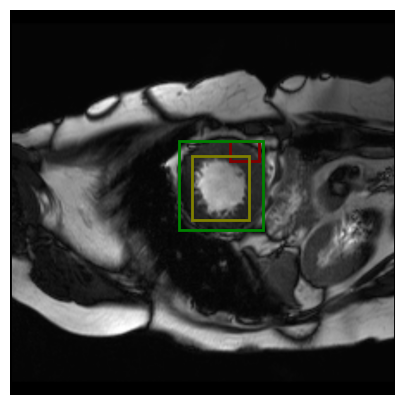}
         \includegraphics[width=\linewidth, trim=100pt 100pt 60pt 60pt, clip]{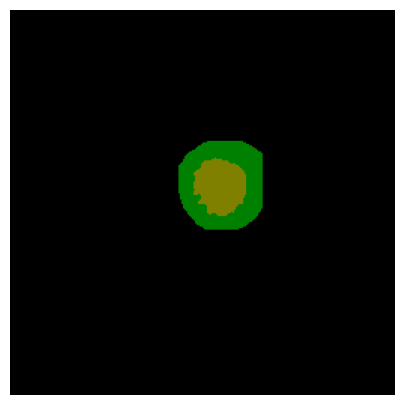}
         \includegraphics[width=\linewidth, trim=40pt 40pt 40pt 40pt, clip]{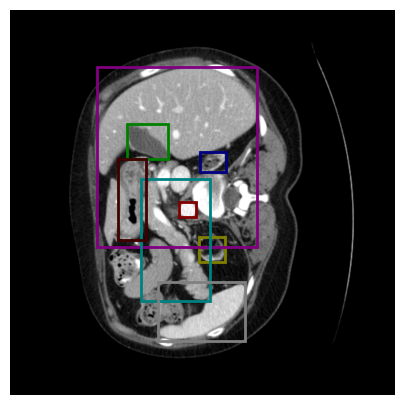}
         \includegraphics[width=\linewidth, trim=40pt 40pt 40pt 40pt, clip]{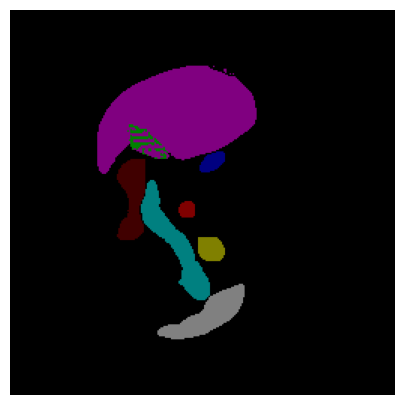}
         \caption{SAM GT Box}
     \end{subfigure}
    \begin{subfigure}[b]{0.195\linewidth}
         \centering
         \includegraphics[width=\linewidth, trim=100pt 100pt 60pt 60pt, clip]{figs/qualitative/sam-box/case_009_volume_ES_7_box.png}
         \includegraphics[width=\linewidth, trim=100pt 100pt 60pt 60pt, clip]{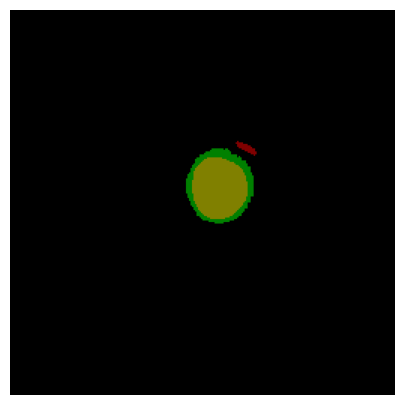}
         \includegraphics[width=\linewidth, trim=40pt 40pt 40pt 40pt, clip]{figs/qualitative/sam-box/case0008_111_box.png}
         \includegraphics[width=\linewidth, trim=40pt 40pt 40pt 40pt, clip]{figs/qualitative/ai-sam-auto/case0008_111_pred.png}
         \caption{MedSAM GT Box}
     \end{subfigure}
        \caption{Qualitative Results of the AI-Prompter on ACDC using 4 points. The first row is the generated prompt and the second row is the corresponding segmentation map. Only the samples that AI-SAM Automatic performs poorly on some classes are selected to visualize how the masks are fixed by the AI-SAM interactive version. The images have been zoomed in to enhance visibility. The segmentation color palette is detailed in Sec.~\ref{sec:metric}.}
        \label{fig:med_qualitative}
\end{figure*}

In this set of experiments, our aim is to evaluate how user prompts affect the output of AI-SAM. We visualize the generated points and the resulting segmentation masks both before and after the addition of a bounding box. We intentionally select the samples that our AI-SAM performs poorly for at least one class under automatic setting to visualize how using the method interactively can correct the segmentation results.

From the medical image segmentation examples in Fig.~\ref{fig:med_qualitative}, we see that the (red class) in ACDC and the (yellow) in Synapse are missing in the automatic setting but are correctly identified with points corrected by the bounding box. However, providing the same bounding box to SAM or MedSAM results in more errors. First, for the class that occupies a smaller part of the bounding box or unintended semantic based on the pre-training data, the class-agnostic prompt of SAM will likely segment the unintended object. The original SAM tends to treat segmentation as a complete part which can be seen from the last row of the segmentation map (pink map) from Fig.~\ref{fig:med_qualitative}; the larger part dominates the smaller part in the bounding box and is ignored. Interactive adaptation can alleviate this problem as shown by the MedSAM GT Box result (pink map) but may introduce additional errors (yellow map). More importantly, we acknowledge that the performance for the class Myo in the ACDC dataset is notably low for both MedSAM and SAM. From Fig.~\ref{fig:med_qualitative}, it is evident that the bounding boxes for Myo and LV often overlap. However, Myo is frequently misclassified as LV. This observation can be explained using the proposed SCM. As illustrated by the upper-right 4 cells in the last SCM of Tab.~\ref{fig:conf_mat}, the bounding box prompt for class Myo (i.e., the second row) exhibits higher FSS than LV (i.e., the first row), making it easy for Myo to be misclassified as LV.

\begin{figure}[t!]
    \centering
     \begin{subfigure}[b]{0.242\linewidth}
         \centering
         \includegraphics[width=\linewidth, trim=20pt 100pt 180pt 100pt, clip]{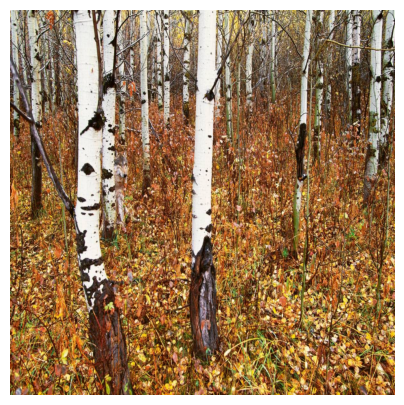}
         \includegraphics[width=\linewidth, trim=20pt 100pt 180pt 100pt, clip]{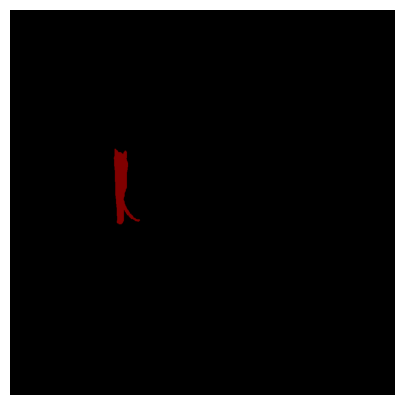}
         \includegraphics[width=\linewidth, trim=20pt 20pt 20pt 20pt, clip]{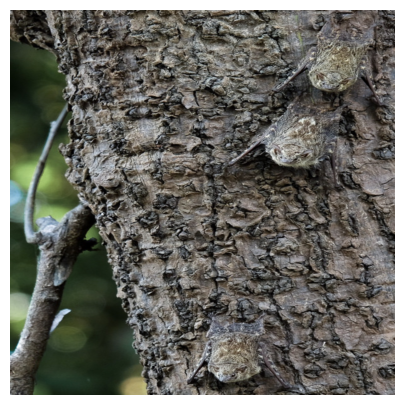}
         \includegraphics[width=\linewidth, trim=20pt 20pt 20pt 20pt, clip]{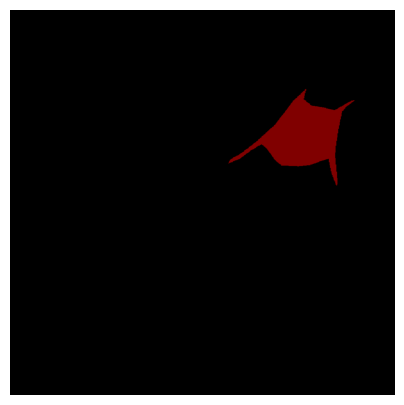}
         \includegraphics[width=\linewidth, trim=20pt 20pt 90pt 90pt, clip]{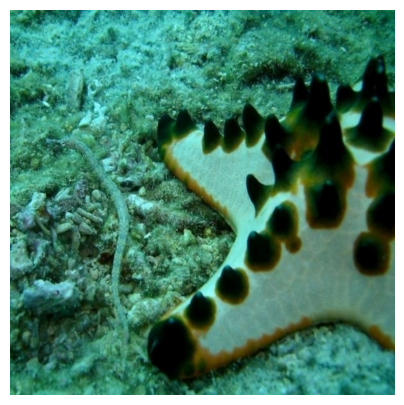}
         \includegraphics[width=\linewidth, trim=20pt 20pt 90pt 90pt, clip]{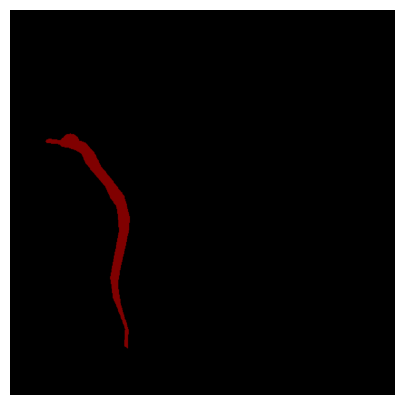}
         \includegraphics[width=\linewidth, trim=70pt 70pt 10pt 10pt, clip]{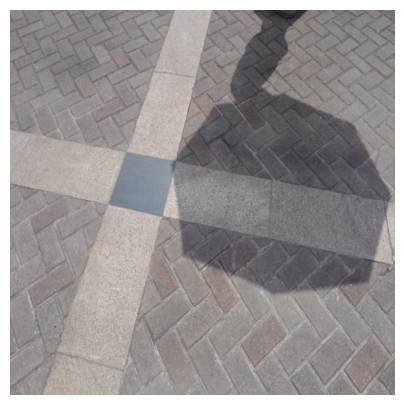}
         \includegraphics[width=\linewidth, trim=70pt 70pt 10pt 10pt, clip]{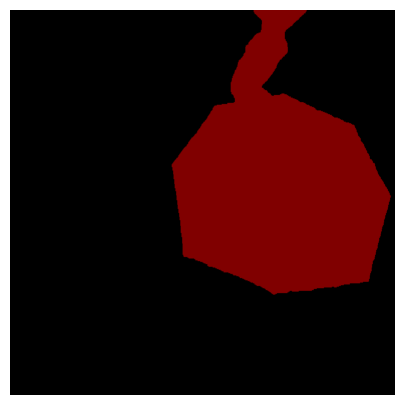}
         \caption{Ground Truth}
     \end{subfigure}
     \begin{subfigure}[b]{0.242\linewidth}
         \centering
         \includegraphics[width=\linewidth, trim=20pt 100pt 180pt 100pt, clip]{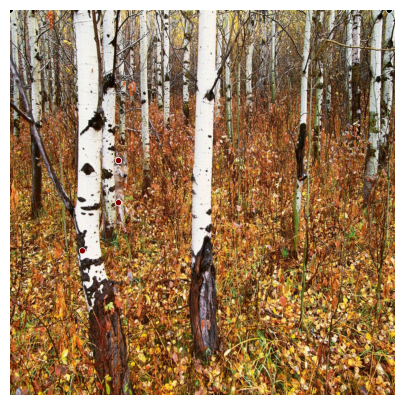}
         \includegraphics[width=\linewidth, trim=20pt 100pt 180pt 100pt, clip]{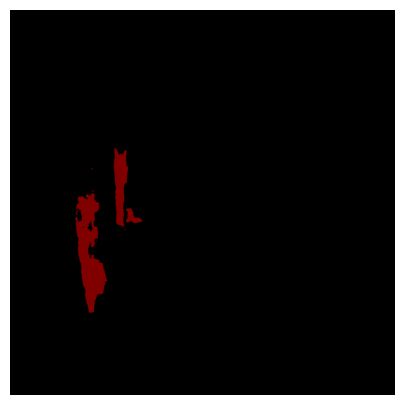}
         \includegraphics[width=\linewidth, trim=20pt 20pt 20pt 20pt, clip]{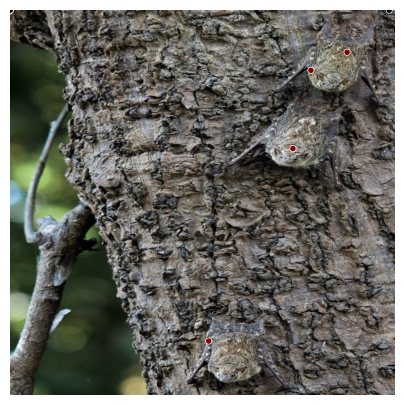}
         \includegraphics[width=\linewidth, trim=20pt 20pt 20pt 20pt, clip]{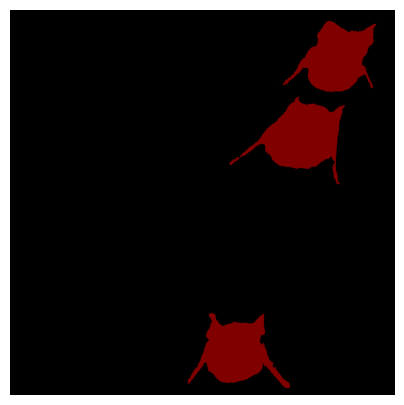}
         \includegraphics[width=\linewidth, trim=20pt 20pt 90pt 90pt, clip]{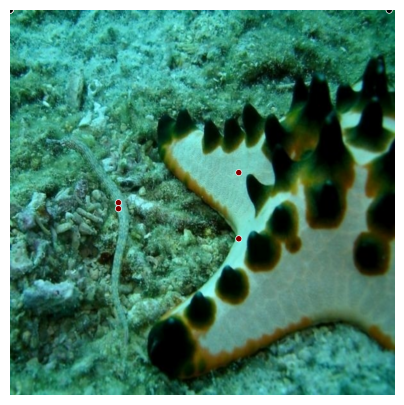}
         \includegraphics[width=\linewidth, trim=20pt 20pt 90pt 90pt, clip]{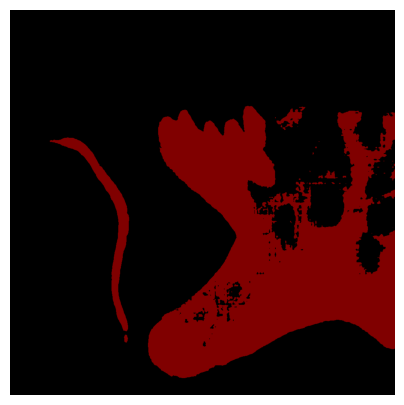}
         \includegraphics[width=\linewidth, trim=70pt 70pt 10pt 10pt, clip]{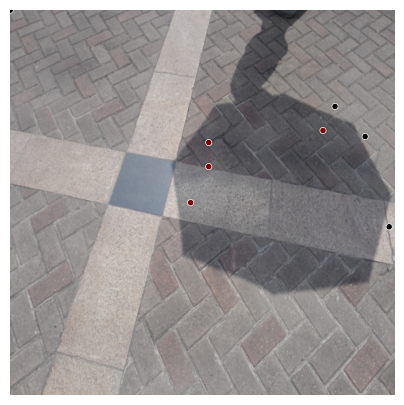}
         \includegraphics[width=\linewidth, trim=70pt 70pt 10pt 10pt, clip]{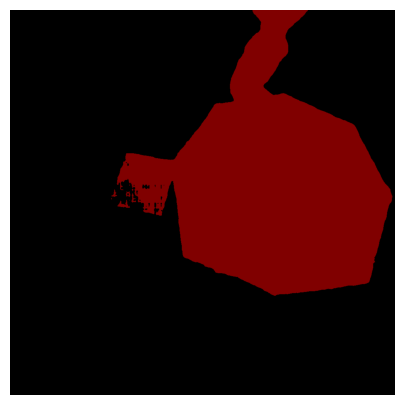}
         \caption{AI-SAM}
     \end{subfigure}
     \begin{subfigure}[b]{0.242\linewidth}
         \centering
         \includegraphics[width=\linewidth, trim=20pt 100pt 180pt 100pt, clip]{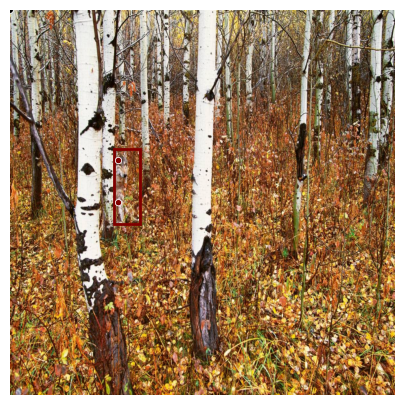}
         \includegraphics[width=\linewidth, trim=20pt 100pt 180pt 100pt, clip]{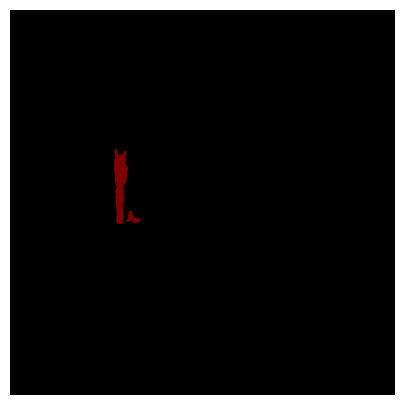}
         \includegraphics[width=\linewidth, trim=20pt 20pt 20pt 20pt, clip]{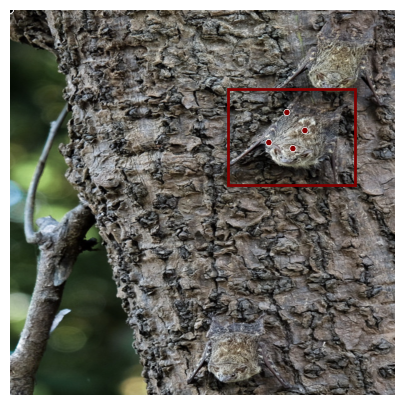}
         \includegraphics[width=\linewidth, trim=20pt 20pt 20pt 20pt, clip]{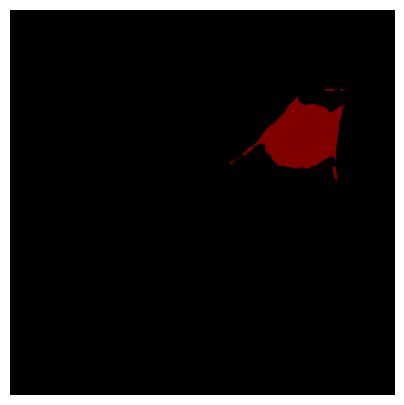}
         \includegraphics[width=\linewidth, trim=20pt 20pt 90pt 90pt, clip]{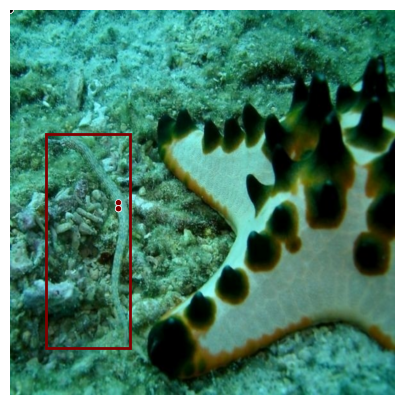}
         \includegraphics[width=\linewidth, trim=20pt 20pt 90pt 90pt, clip]{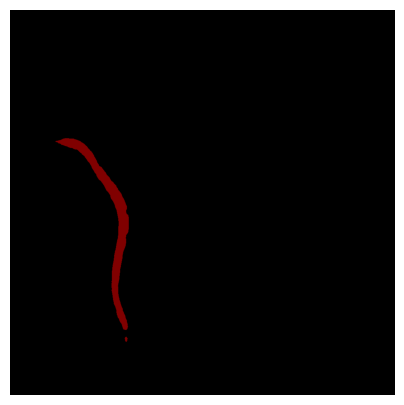}
         \includegraphics[width=\linewidth, trim=70pt 70pt 10pt 10pt, clip]{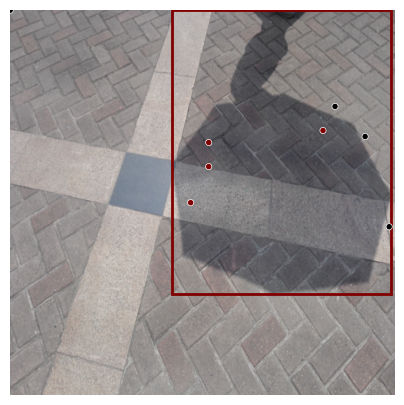}
         \includegraphics[width=\linewidth, trim=70pt 70pt 10pt 10pt, clip]{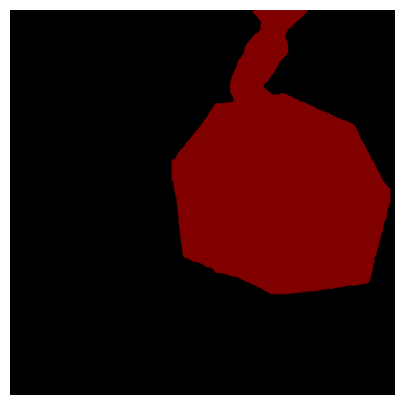}
         \caption{AI-SAM Box}
     \end{subfigure}
    \begin{subfigure}[b]{0.242\linewidth}
         \centering
         \includegraphics[width=\linewidth, trim=20pt 100pt 180pt 100pt, clip]{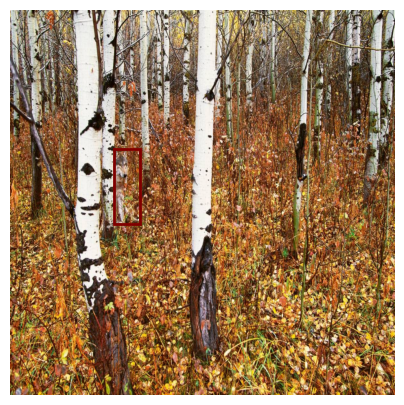}
         \includegraphics[width=\linewidth, trim=20pt 100pt 180pt 100pt, clip]{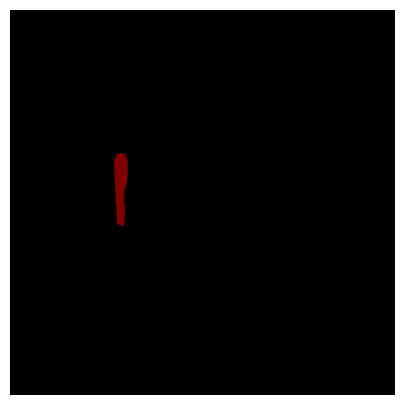}
         \includegraphics[width=\linewidth, trim=20pt 20pt 20pt 20pt, clip]{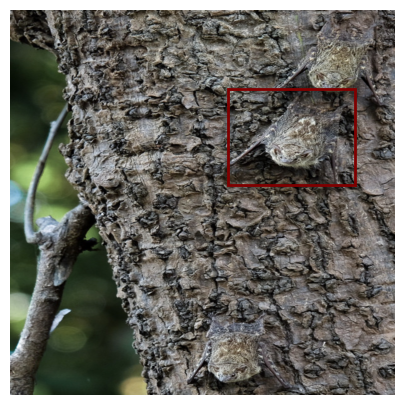}
         \includegraphics[width=\linewidth, trim=20pt 20pt 20pt 20pt, clip]{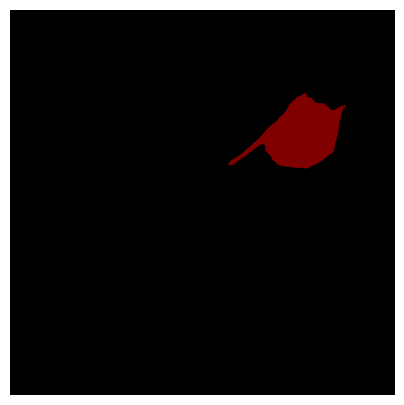}
         \includegraphics[width=\linewidth, trim=20pt 20pt 90pt 90pt, clip]{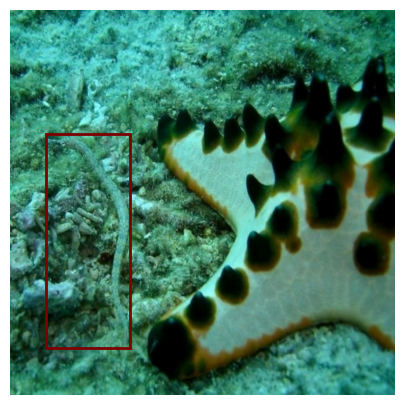}
         \includegraphics[width=\linewidth, trim=20pt 20pt 90pt 90pt, clip]{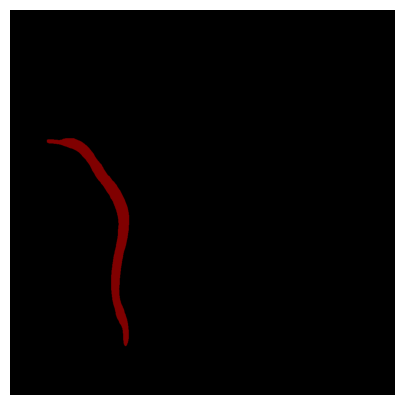}
         \includegraphics[width=\linewidth, trim=70pt 70pt 10pt 10pt, clip]{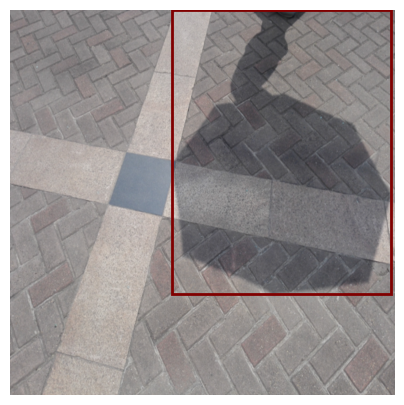}
         \includegraphics[width=\linewidth, trim=70pt 70pt 10pt 10pt, clip]{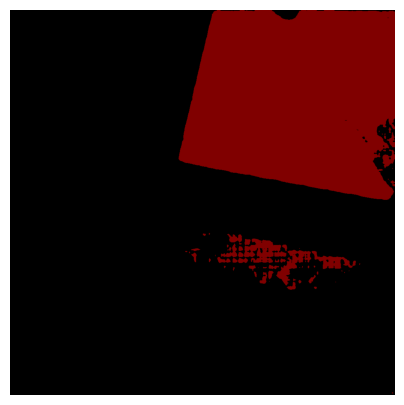}
         \caption{SAM Box}
     \end{subfigure}
        \caption{Qualitative Results of the AI-SAM on COD tasks and shadow detection task using 4 points. The first row is the generated prompt and the second row is the corresponding segmentation map. Only the samples that AI-SAM Automatic performs poorly are selected to visualize how the masks are fixed by the AI-SAM interactive version. The images have been zoomed in to enhance visibility.}
        \label{fig:natural_qualitative}
\end{figure}

Next, in the automatic camouflage detection and shadow detection task shown in Fig.~\ref{fig:natural_qualitative}, we see that AI-SAM tends to identify all the objects in one image as shown by the generated points as well as the segmentation masks, which results in a large disagreement with the ground truth label. Moreover, in shadow detection, AI-SAM tends to also treat the darker part that is connected to the shadow as part of the shadow. All of the above errors can be fixed by introducing additional user prompts which highlight the superiority of an automatic and interactive method. We note that SAM with a bounding box prompt can already generate good results in those cases but in general, has less detail or can provide a very inaccurate result as shown by the shadow detection example in the last row of Fig.~\ref{fig:natural_qualitative}.

\subsection{Ablation Study}
\noindent\textbf{Number of Points.} To investigate how the number of points affects the model's performance, we conducted experiments using different numbers of points on the ACDC dataset. As shown in Table~\ref{tab:num_points}, the performance remains relatively consistent. However, as we use too many points, the performance begins to degrade. Thus, the number of points is not a contributing factor as long as we use a reasonable number of points.

At first glance, this result may seem counterintuitive, as one might expect better performance with more information provided. However, there are key factors at play. SAM training typically utilizes up to 16 points in total to generate a mask. In our approach, we utilize the foreground points of other classes as background points for the current class. For instance, in the ACDC dataset with 4 classes, including the background, if we have 16 points for each class, we would generate a total of 64 points. This number significantly exceeds the number of pre-training points, effectively extrapolating the mask decoder and leading to a performance decline. Additionally, learning to generate more points is a more challenging task, which can result in fewer optimal points.

\begin{table}[h!]
    \centering
    \begin{tabular}{c|ccccc}
        \toprule
        \# Points& 1&2&4&8&16 \\
        \midrule
        DICE& 92.17& 92.25& 92.06& 92.05& 91.77\\
        \bottomrule
    \end{tabular}
    \caption{The ablation study on the number of point prompts per class. Original SAM uses a maximum of 16 points which is equivalent to 4 points per class in our setting.}
    \label{tab:num_points}
\end{table}

\noindent\textbf{Prompt Heuristic Loss.}
The design of the loss functions aims to align the generated prompts with the intuitions of usability presented in Sec.~\ref{sec:phl} while maintaining good model performance. To evaluate the effectiveness of the Prompt Heuristic Loss design, we employ both qualitative and quantitative criteria for assessment.

In qualitative evaluation, we visually inspect the generated prompts to ensure that the points are located on the object of interest and that the correct number of points are generated. For quantitative evaluation, we assess the impact of the proposed losses on model performance by conducting ablation studies. Specifically, we investigate the effects of removing individual components of the loss design and transitioning from generalized points to one-hot points.

Our ablation studies include the following scenarios: removing the diversity loss $\mathcal{L}^{pd}$, removing both the diversity loss and sharpness loss $\mathcal{L}^{ps}$, and removing the entire prompt heuristic loss $\mathcal{L}^{ph}$.

The results presented in Tab.~\ref{tab:loss_abl} highlight the significance of these loss functions. Notably, the removal of the diversity loss $\mathcal{L}^{pd}$ does not significantly affect the model's ability to utilize one-hot points, as expected. However, when we remove the entire prompt heuristic loss $\mathcal{L}^{ph}$, a substantial drop in performance (i.e., 41.32) is observed, underscoring the critical importance of the prompt heuristic loss for usability.
\begin{table}[h!]
    \centering
    \begin{tabular}{l|cc|c}
        \toprule
        Point Type & $P^g$ & $P$ & $\Delta$ \\
        \midrule
        AI-SAM & 92.06 &  92.06 & 0.0\\
        w/o $\mathcal{L}^{pd}$ & 92.12 &  92.12 & 0.0\\
        w/o $\mathcal{L}^{pd},\mathcal{L}^{ps}$ & 91.99 &  91.98 & 0.01\\
        w/o $\mathcal{L}^{ph}$ & 91.75 & 50.43 & 41.32\\
        \bottomrule
    \end{tabular}
    \caption{Ablation Study on ACDC using AI-Prompter with 4 points. Mean DICE scores (\%) are used. $\Delta$ is the change in performance by switching point type from generalized point $P^g$ to one-hot point $P$.}
    \label{tab:loss_abl}
\end{table}

\begin{figure}[h!]
     \centering
     \begin{subfigure}[b]{0.24\linewidth}
         \centering
         \includegraphics[width=\linewidth, trim=40pt 50pt 40pt 30pt, clip]{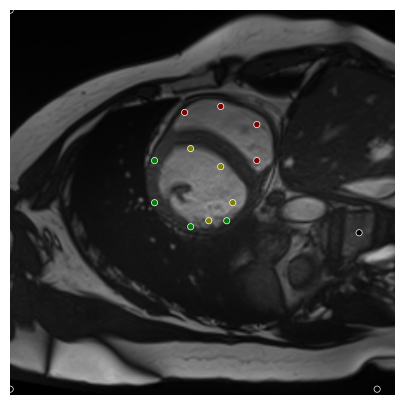}
         \includegraphics[width=\linewidth, trim=40pt 50pt 40pt 30pt, clip]{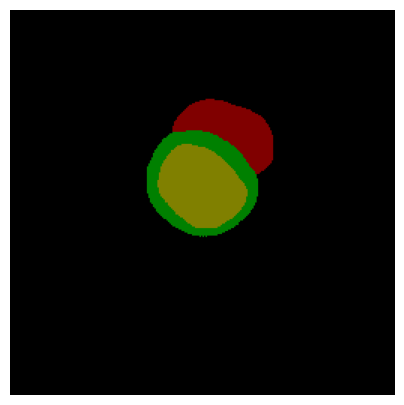}
         \caption{AI-Prompter}
         \label{fig:acdc_abl_all}
     \end{subfigure}
     \begin{subfigure}[b]{0.24\linewidth}
         \centering
         \includegraphics[width=\linewidth, trim=40pt 50pt 40pt 30pt, clip]{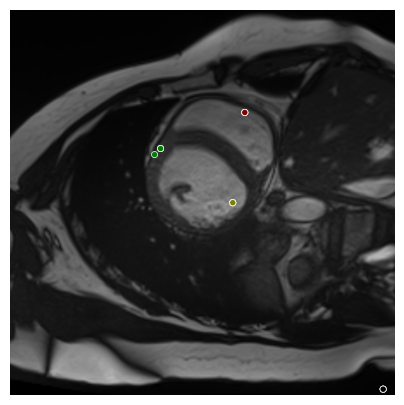}
         \includegraphics[width=\linewidth, trim=40pt 50pt 40pt 30pt, clip]{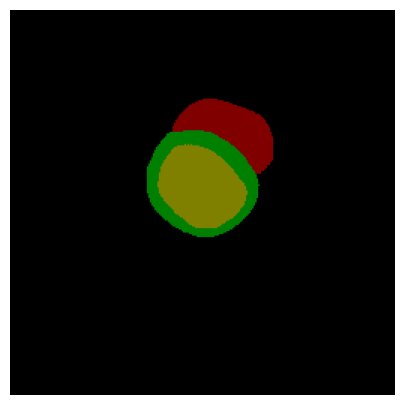}
         \caption{w/o $\mathcal{L}^{pd}$}
         \label{fig:acdc_abl_all_no_div}
     \end{subfigure}
     \begin{subfigure}[b]{0.24\linewidth}
         \centering
         \includegraphics[width=\linewidth, trim=40pt 50pt 40pt 30pt, clip]{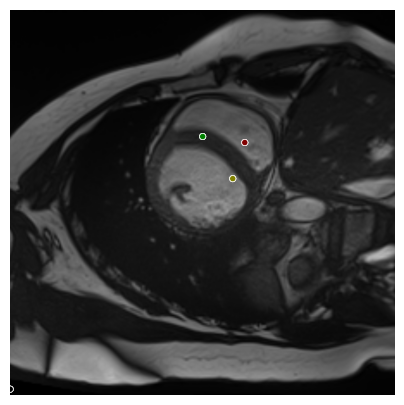}
         \includegraphics[width=\linewidth, trim=40pt 50pt 40pt 30pt, clip]{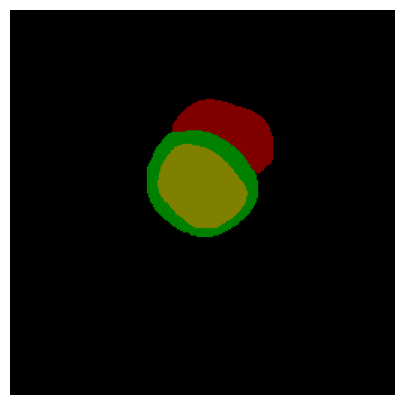}
         \caption{w/o $\mathcal{L}^{pd},\mathcal{L}^{ps}$}
         \label{fig:acdc_abl_all_no_div_sharp}
     \end{subfigure}
    \begin{subfigure}[b]{0.24\linewidth}
         \centering
         \includegraphics[width=\linewidth, trim=40pt 50pt 40pt 30pt, clip]{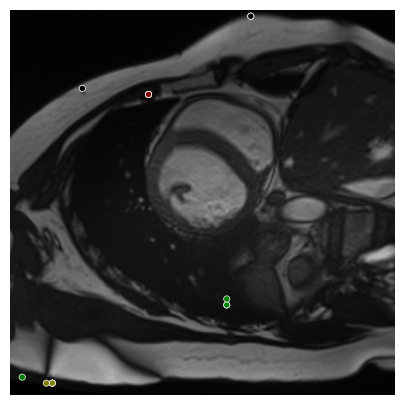}
         \includegraphics[width=\linewidth, trim=40pt 50pt 40pt 30pt, clip]{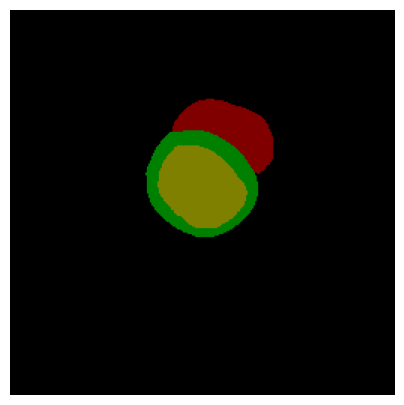}
         \caption{w/o $\mathcal{L}^{ph}$}
         \label{fig:acdc_abl_all_no_ph}
     \end{subfigure}
        \caption{Qualitative Results of the AI-Prompter on ACDC using 4 points. The first row is the generated prompt and the second row is the corresponding segmentation map. The images have been zoomed in to enhance visibility. The segmentation color palette is detailed in Sec.~\ref{sec:metric}.}
        \label{fig:acdc_abl}
\end{figure}

When comparing the results of AI-SAM with and without $\mathcal{L}^{ph}$ (92.12 vs. the results in Fig.~\ref{tab:num_points}), it is likely that the model produces fewer points as its performance falls within that range. To further examine the effect of the diversity loss, we visually represented the generated points in Fig.~\ref{fig:acdc_abl}. When the complete prompt heuristic loss is applied, we obtain four points for each class, and these points align with human intuitions. However, upon removing the diversity loss, we observe that almost all points converge to the same location, which is consistent with our intuition in Sec.~\ref{sec:phl} that encourages points to be similar, as supported by the qualitative results in Fig.~\ref{fig:acdc_abl_all_no_div}.

Furthermore, if we additionally eliminate the point sparsity loss, the model not only exhibits behavior similar to that when the diversity loss is removed (as shown in Fig.~\ref{fig:acdc_abl_all_no_div_sharp}) but also performs worse. Furthermore, we start to see differences between generalized points and one-hot points (as indicated in \ref{tab:loss_abl}). Lastly, when the entire prompt heuristic loss is removed, the model not only produces fewer than four points for each class but also generates many incorrect points (as shown in Fig.~\ref{fig:acdc_abl_all_no_ph}), which aligns with our intuition that a neural network can learn class representation instead point locations without proper supervision. These ablation results collectively underscore the effectiveness of each proposed loss in serving its intended purpose of improving usability.

\noindent\textbf{Prompt Quality.} 
While we have demonstrated that adding a point sampled from the ground truth segmentation mask can enhance model performance (as shown in Table~\ref{tab:medical_sotas}) and visualized the points generated by AI-Prompter (as shown in Fig.~\ref{fig:acdc_abl}), the superiority of the AI-Prompter-generated points over those sampled from the ground truth segmentation mask remains to be established. Although the proposed PCM could potentially assess prompt quality, our current implementation, as outlined in Sec.~\ref{sec:property_prompt}, does not fulfill this purpose. To empirically validate the quality of the points generated by AI-Prompter, we conducted an experiment where we replaced the AI-Prompter-generated points with points sampled from the ground truth on the ACDC dataset. The result was a significant decrease in the DICE score, from 92.06 to 64.05. This finding underscores the effectiveness of AI-Prompter in generating high-quality prompts.

\section{Discussion}

\subsection{Limitations}
In the development of our method, we encountered certain limitations of the current SAM and interactive segmentation paradigm. First, a key challenge lies in the misalignment between the class semantics of the pre-training data and the downstream dataset. This issue becomes particularly evident in applications such as shadow detection, as also discussed in a previous study~\cite{chen2023sam}. SAM is primarily trained on natural images where shadow labels are often neglected. Therefore, applying SAM directly to downstream tasks can result in subpar performance. Additionally, SAM encourages the use of bounding box prompts alongside point prompts. However, in our work, we have identified that using bounding boxes alone as prompts can introduce ambiguity, as demonstrated by our PCM in Sec.~\ref{sec:property_prompt}. While selecting a suitable point is indeed a challenging task, it is essential to reduce ambiguity, and this is a key contribution of our proposed approach.

Furthermore, our method also has its own set of limitations. Firstly, the effectiveness of our proposed points relies on the representational power of the image encoder and decoder to produce high-quality segmentation results. Additionally, to generate class-specific points, the class label must be provided in advance. Unless fine-tuning is performed, AI-SAM cannot be used automatically for new classes. One potential avenue for addressing this limitation is the application of open-vocabulary or vision-language segmentation methods to enable automatic adaptation to any class labels. Moreover, applying foreground and background segmentation, where all classes are treated as foreground, may hinder the ability of the current AI-Prompter to generate background points. This is because we currently use the foreground points from other classes as background points. This issue could be mitigated by learning background points explicitly.

\subsection{Broader Implications}

An automatic and interactive method can greatly benefit model development in various settings, particularly in cases where training data is limited in quantity or diversity. For instance, in autonomous driving applications, engineers often collect and label natural images under different road, weather, and lighting conditions. An automatic and interactive method can streamline this process by automatically labeling straightforward samples and requiring only a few clicks for challenging samples, thus reducing the burden of dense annotations.

In medical applications, expert knowledge is essential for accurate labeling. By employing an automatic and interactive method, we can significantly reduce the labeling time. This reduction occurs by eliminating the time experts spend on simpler samples and gradually reducing their involvement as the model undergoes iterative re-training on an increasing amount of annotated data.

Moreover, there are applications that fall between natural and medical image domains, such as photographic-based placenta analysis. The placenta serves as a window to assess pregnancy outcomes and various health conditions in newborns. Unfortunately, only a fraction of placentae are examined by pathologists due to the associated examination costs. Previous work has demonstrated that deep learning methods can predict many outcomes using photographic images~\cite{chen2020ai,pan2022vision,pan2023enhancing}. However, the main challenge lies in the dependency on high-quality segmentation masks. This dependency arises from biases in data collection, where training data predominantly originates from a single site or a few sites, resulting in higher quality and more uniform images. In contrast, testing data may exhibit diverse backgrounds, variable bloodstain distributions around the disk, and different types of rulers, as highlighted by Zhang et al.~\cite{zhang2020multi}. These data biases can lead to underperforming models. AI-SAM can seamlessly address this issue by allowing users to easily modify segmentation maps to correct errors and improve model performance.

\section{Conclusion}
We introduced AI-SAM, a novel paradigm that bridges the gap between automatic and interactive segmentation. We analyzed different prompt types and provided a method to generate effective prompts. This unified framework not only introduces a new approach to segmentation but also holds significant promise in real-time medical imaging applications. We anticipate that AI-SAM will inspire further advancements in the field.

%% file: sec/X_suppl.tex

\section{Implementation Details}
\label{sec:rationale}
\subsection{Training}
In addition to supervising AI-Prompter using the prompt heuristic loss $\mathcal{L}^{ph}$, we follow the supervision approach of previous automatic adaptation methods for AI-SAM. Specifically, we apply the cross-entropy loss and DICE loss for medical image segmentation, use binary cross-entropy (BCE) loss and intersection over union loss for the COD setting, and employ BCE loss for shadow detection.

\subsection{Architectures}

\noindent\textbf{Classifier.} The classifier's objective is to conduct multi-class classification on a given image to filter out points associated with non-existing objects. In principle, any classifier can be employed. In our implementation, we draw inspiration from Query2label~\cite{liu2021query2label} and utilize a transformer decoder to generate a list of tokens for all the classes, subsequently performing binary classification on each class token. We employ the ASL loss~\cite{ridnik2021asymmetric} to train the model. It's worth noting that using a transformer encoder, as done in Query2label, would be highly inefficient, as it would require computing a 4096$\times$4096 matrix. Instead, we leverage our attention-convolution design from AI-Prompter to aggregate image features without the need for a transformer encoder. The architecture of the classifier block is depicted in Fig.\ref{fig:ai-prompter}, with the exclusion of the final softmax layer.
\begin{figure}[h!]
    \centering
    \includegraphics[width=\linewidth]{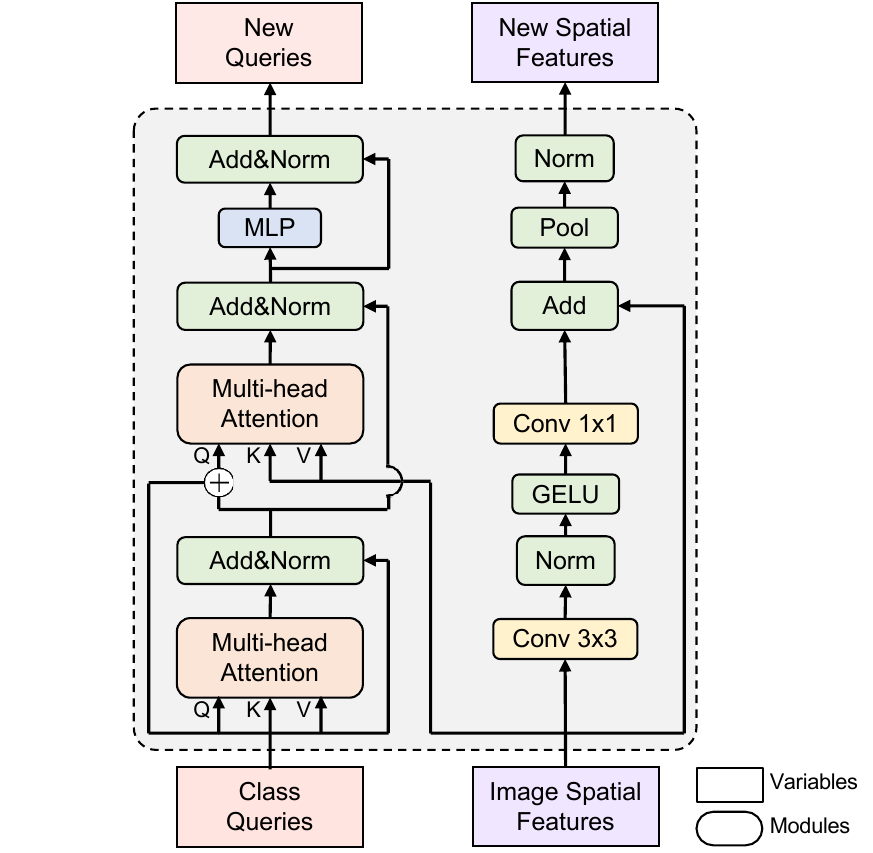}
    \caption{The illustration of the classifier block.}
    \label{fig:classifier-block}
\end{figure}

\subsection{Hyper-parameters}
All experiments were conducted using 4 NVIDIA A40-48GB GPUs. We utilized the vit\_h version as the base SAM model and adopted the LoRA adaptation method for the image encoder, similar to SAMed\_h, to conserve computational resources. The prompt encoder was kept fixed, and we trained the AI-Prompter, classifier, and mask decoder jointly with the image encoder. The classifier was trained using the ASL loss. Since our primary objective was not to maximize automatic segmentation performance, we performed minimal hyperparameter selection. Specifically, for the medical image segmentation experiments (i.e., ACDC and Synapse), we used the exact same hyperparameters as presented in Table~\ref{tab:hyper-parameters}. This approach contrasts with previous methods where parameters were fine-tuned for each dataset. For camouflage and shadow detection, we adopted the exact hyperparameters from SAM-Adapter~\cite{chen2023sam}. However, we employed checkpointing~\footnote{https://pytorch.org/docs/stable/checkpoint.html} on the encoder and adjusted the batch size to 16 to fully utilize the available GPU memory, as our GPUs had less VRAM compared to SAM-Adapter.
\begin{table}[ht]
    \centering
    \begin{adjustbox}{width=\linewidth}
    \begin{tabular}{l|c}
        \toprule
        Name & Value\\
        \midrule
        Scheduler & CosineLRScheduler\\
        Base LR. &  1e-4 \\
        Total Epochs & 80 \\
        Warmup Epochs & 8 \\
        Warmup Init. LR. & 1e-10 \\
        \midrule
        Optimizer & AdamW \\
        Betas & (0.9, 0.999) \\
        Eps & 1e-08 \\
        Weight Decay & 1e-2\\
        \midrule
        Cross-entropy Loss Weight & 0.3 \\
        DICE Loss Weight & 0.7 \\
        Diversity Loss Temp. & 7 \\
        Prompt Heuristic Loss Weight $\gamma$ & 0.1\\
        Correctness Loss Weight $\alpha^{pc}$ & 2.0 \\
        Sharpness Loss Weight $\alpha^{ps}$ & 1.0 \\
        Intra-class Diversity Weight $\beta^{out}$ & 0.5 \\
        Inter-class Diversity Weight $\beta^{in}$ & 0.2 \\
        \midrule
        ASL Loss Weight & 1.0 \\
        ASL Gamma+ & 0.0 \\
        ASL Gamma- & 2.0 \\
        \bottomrule
    \end{tabular}
    \end{adjustbox}
    \caption{AI-SAM Hyper-parameters for medical image segmentation.}
    \label{tab:hyper-parameters}
\end{table}

\subsection{Preprocessing}
 All inputs were reshaped to 1024$\times$1024 and normalized following the original SAM protocol. In the case of medical image segmentation, we employed the same preprocessing methods as in previous work. To enhance training stability, we incorporated random cropping as an additional data augmentation technique, especially since SAMed encountered instability issues during the training of ViT\_h. For COD and Shadow detection, we follow the exact preprocessing methods in SAM-Adapter.

\subsection{Evaluation Metric}
\label{sec:metric}
The performance evaluation of AI-SAM on each task follows previous works. The color pallette for the segmentation masks is shown in Table~\ref{tab:color_pallete}. 

\noindent\textbf{Synapse Multi-organ dataset.} Performance is evaluated using DICE score and $95\%$ Hausdorff Distance (HD95).

\noindent\textbf{Automated Cardiac Diagnosis Challenge (ACDC).} Performance is evaluated using DICE score.

\noindent\textbf{Camouflaged Object Detection.} Performance is evaluated using commonly used metrics: S-measure ($S_\alpha$), mean E-measure ($E_\phi$), weighted F-measure ($F^\omega_\beta$ ), and MAE for evaluation.

\noindent\textbf{Image Shadow Triplets Dataset (ISTD).} Performance isevaluated  using balance error rate (BER).

\begin{table}[h!]
    \centering
    \begin{adjustbox}{width=\linewidth}
    \begin{tabular}{l|c|c|c|c|c|c|c|c|c}
    \toprule
         Dataset & \multicolumn{9}{c}{Class} \\
         \midrule
         \multirow{2}{*}{ACDC} & None & RV & Myo & LV\\
         &\cellcolor[rgb]{0. 0. 0.} & \cellcolor[rgb]{0.5 0. 0.}& \cellcolor[rgb]{0. 0.5 0.}& \cellcolor[rgb]{0.5 0.5 0.} \\
         \midrule
         \multirow{2}{*}{Synapse} & None & Aorta & GB & KL & KR & Liver & PC & SP & SM \\
             &\cellcolor[rgb]{0. 0. 0.} & \cellcolor[rgb]{0.5 0. 0.}& \cellcolor[rgb]{0. 0.5 0.}& \cellcolor[rgb]{0.5 0.5 0.} & \cellcolor[rgb]{0. 0. 0.5} & \cellcolor[rgb]{0.5 0. 0.5}& \cellcolor[rgb]{0. 0.5 0.5}& \cellcolor[rgb]{0.5 0.5 0.5} & \cellcolor[rgb]{0.25 0. 0.}\\
             \bottomrule
    \end{tabular}
    \end{adjustbox}
    \caption{The color palette for medical image segmentation datasets.}
    \label{tab:color_pallete}
\end{table}